\documentclass[10pt,twocolumn,letterpaper]{article}
\usepackage[dvipsnames,table]{xcolor}
\usepackage[pagenumbers]{cvpr} %

\newcommand{\duster}[0]{DUSt3R}

\newcommand{\method}[0]{\textcolor{black}{InterPose}}

\usepackage{multirow}
\usepackage{booktabs}
\usepackage{tabularx}
\usepackage{graphicx}
\usepackage{gensymb}
\usepackage{caption}
\definecolor{yellow}{rgb}{1, 1, 0.7}
\definecolor{orange}{rgb}{1, 0.85, 0.7}
\definecolor{red}{rgb}{1, 0.7, 0.7}

\definecolor{cvprblue}{rgb}{0.21,0.49,0.74}
\usepackage[pagebackref,breaklinks,colorlinks,allcolors=cvprblue]{hyperref}

\usepackage[capitalize]{cleveref}
\crefname{section}{Sec.}{Secs.}
\Crefname{section}{Section}{Sections}
\Crefname{table}{Table}{Tables}
\crefname{table}{Tab.}{Tabs.}
\Crefname{equation}{Equation}{Equations}
\crefname{equation}{eq.}{eqs.}

\def\paperID{****} %
\def\confName{CVPR}
\def\confYear{2025}

\title{Can Generative Video Models Help Pose Estimation?}

\author{Ruojin Cai$^{1,2}$
\quad
Jason Y. Zhang$^{1}$
\quad
Philipp Henzler$^{1}$
\quad
Zhengqi Li$^{1}$\\
\quad
Noah Snavely$^{1,2}$
\quad
Ricardo Martin-Brualla$^{1}$
\vspace{2mm}
\\
\centerline{$^1$Google \quad $^2$Cornell University}
}

\begin{document}

\twocolumn[{%
\maketitle
\thispagestyle{empty}
\begin{center}
\centering
    \includegraphics[width=\linewidth, trim=0 15 45 0, clip]{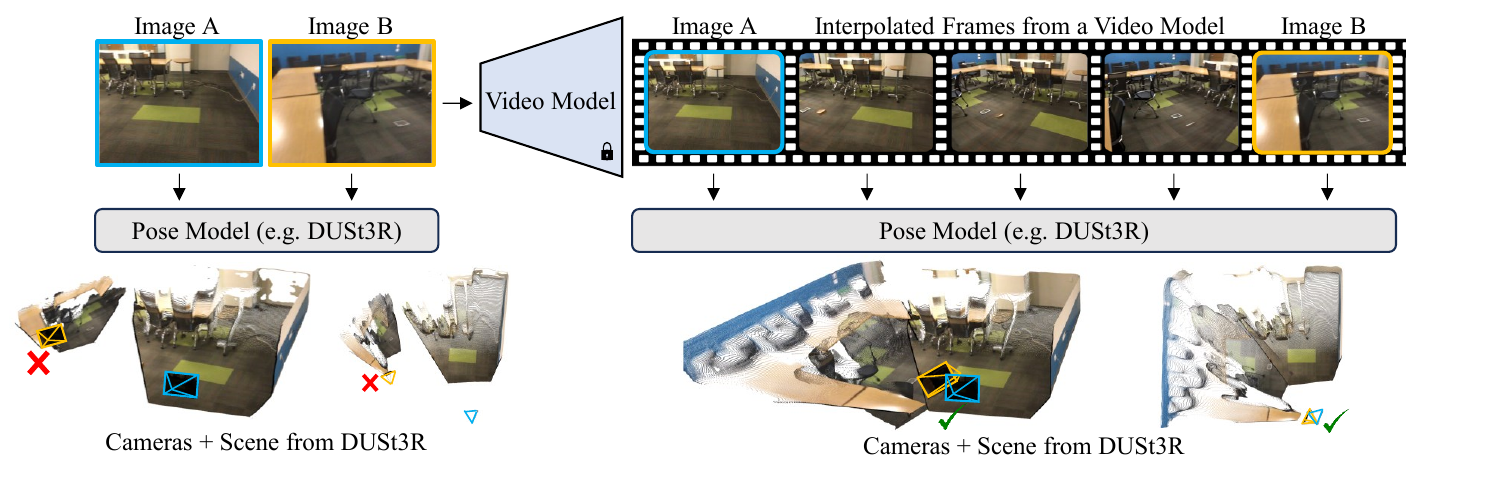}
\captionof{figure}{\textbf{Improving pose estimation by interpolating frames using a video model.} Given two images of a scene with almost no overlap, we aim to recover their relative camera pose. Without being able to rely on visual correspondences, existing methods struggle in this setting (left). We propose to use an off-the-shelf video generation model to interpolate a video connecting the two images. 
    Augmented with the frames generated by the video model, existing pose estimators (e.g. \duster~\cite{wang2024dust3r}) are able to more accurately recover the correct pose (right).
}%
\label{fig:teaser-fig}%
\end{center}%
}]

\begin{abstract}

Pairwise pose estimation from images with little or no overlap is an open challenge in computer vision. Existing methods, even those trained on large-scale datasets, struggle in these scenarios due to the lack of identifiable correspondences or visual overlap.  Inspired by the human ability to infer spatial relationships from diverse scenes, we propose a novel approach, \method, that leverages the rich priors encoded within pre-trained generative video models. We propose to use a video model to hallucinate intermediate frames between two input images, effectively creating a dense, visual transition, which significantly simplifies the problem of pose estimation.
Since current video models can still produce implausible motion or inconsistent geometry, we introduce a self-consistency score that evaluates the consistency of pose predictions from sampled videos.
We demonstrate that our approach generalizes among three state-of-the-art video models and show consistent improvements over the state-of-the-art \duster~baseline on four diverse datasets encompassing indoor, outdoor, and object-centric scenes.
Our findings suggest a promising avenue for improving pose estimation models by leveraging large generative models trained on vast amounts of video data, which is more readily available than 3D data. See our project page for results: \url{Inter-Pose.github.io}.

\end{abstract}

\section{Introduction}
\label{sec:intro}

Consider the classroom in \cref{fig:teaser-fig}. We, as humans, can %
reasonably guess the spatial relationship between the two images, recognizing that the table on the left side of the first image is the same as the table on the right side of the second image.  Even though the images are taken from %
viewpoints with almost no overlap, we leverage our prior knowledge about typical classroom layouts to infer this connection. This task of determining the relative pose between two images is a core component of all pose estimation pipelines and a pre-requisite for most tasks in 3D computer vision.

Traditional approaches to pairwise pose estimation rely on identifying and matching features between an image pair~\cite{lowe2004distinctive} to compute the relative geometric transformation~\cite{hartley1997defense}. While effective when images have significant overlap and textural details, these methods struggle when faced with drastically different viewpoints, as seen in our classroom example.
Recent advances in deep learning have led to more robust pose estimators. The groundbreaking \duster~\cite{wang2024dust3r} model is trained on a mixture of several large-scale 3D datasets, and demonstrates impressive performance and generalization ability. However, even such a sophisticated method struggles with extreme viewpoint changes where establishing correspondences becomes impossible.

Unlike 3D understanding models like \duster, video models can be pre-trained on vast amounts of web-scale video data,  orders of magnitude larger than 3D datasets. The scale of the data allows for training models that learn significantly more powerful priors of the visual world compared to 3D understanding models. For instance, state-of-the-art video models can generate videos with complex camera motions moving through a scene, reflections on shiny materials, and dynamic subjects undergoing complex interactions, and can be prompted by images or text. Our goal is to tap into this extracted knowledge for downstream scene understanding tasks, like pose estimation.

An exciting application of such generative video models is to generate videos that interpolate between two given key frames~\cite{xing2025dynamicrafter}. Thanks to the learned visual prior, the generated interpolated videos can display plausible, 3D consistent camera motions that transform one video into another. We observe that such hallucinations are providing an explanation of the scene, and in turn, we can use those hallucinated frames to better understand the scene. In this paper, we propose \method, which demonstrates that feeding generated interpolated frames along with the original input pair to state-of-the art pose estimation methods can improve their robustness and accuracy over using the original pair alone.

In some cases, generated videos may contain visual inconsistencies, like morphing or shot cuts, that can degrade pose estimation performance. One approach is to sample multiple such video interpolations, with the hope that one displays a plausible interpretation of the scene that is 3D consistent. However, how do we tell which video sample is a good one? 

We address this by introducing a self-consistency score to evaluate the reliability of the predicted pose for a given video.
Our method samples different sets of frame indices from the interpolated video, and computes multiple pose estimates using these frames together with the input image pair, creating multiple pose estimates per sampled video.
An ideal pose prediction comes from a video whose pose estimates are invariant to the specific sampled frame indices, e.g., whose pose estimates are tightly clustered, and among the pose estimates from that video, one that is close to the other estimates, e.g., the centroid or medoid.

Although simple, we demonstrate the efficacy of our method on challenging input pairs extracted from four diverse datasets, including indoor, outdoor and object-centric scenes. 
In summary, our key contributions include:
\begin{itemize}
    \item we demonstrate for the first time that a generative video model can improve pose estimation by acting as a world prior, improving on the results of a state-of-the-art pose estimator (DUSt3R);
    \item we present a new benchmark of challenging image pairs with small to no overlap across four different datasets encompassing outdoor scenes, indoor scenes, and object-centric views;
    \item and we propose a simple-yet-effective way to score the self-consistency of estimated poses from interpolated videos that generalizes across three different publicly available video models.
\end{itemize}

\section{Related work}
\label{sec:related}
\subsection{Generative Video Models}

Early efforts to build video generators based on GANs~\cite{vondrick2016generating,tulyakov2017mocogan,saito2017temporal,lee2018stochastic} and VAEs~\cite{denton2018stochastic,hsieh2018learning,villegas2018hierarchical} had limited visual fidelity.
More recently, diffusion models~\cite{ho2020denoising,sohl2015deep,song2020denoising} have revolutionized generative image \cite{rombach2021high,ramesh2022hierarchical,saharia2022photorealistic} and video generation.
Earlier diffusion-based models often made predictions directly in pixel space~\cite{ho2022video,ho2022imagen,singer2022make}. Such architectures made it computationally expensive to predict high resolution image frames. To alleviate this issue, subsequent works looked at making predictions in the latent space of an autoencoder~\cite{Gupta2023PhotorealisticVG,bar2024lumiere,villegas2022phenaki,blattmann2023align,xing2025dynamicrafter}.
Since then a variety of video models has been released that demonstrates near-photorealism at high resolution. These models are only available behind a paywall~\cite{lumaDreamMachine,runway,kling} or are not available to the public at all \cite{brooks2024sora}. In our work, we evaluate both public and commercial video models.

\subsection{Relative Pose Estimation}

The classic approach to computing the pose between two images is to extract image features~\cite{lowe2004distinctive,bay2006surf,rublee2011orb}, find correspondences~\cite{muja2009fast}, and then compute the fundamental matrix~\cite{hartley1997defense,longuet1981computer,nister2004efficient} while rejecting outliers~\cite{fischler1981random}. Learning-based methods have significantly improved each of these components, providing better features~\cite{detone2018superpoint,tyszkiewicz2020disk} and matchers~\cite{sarlin2020superglue,lindenberger2023lightglue,jiang2024Omniglue,karpur2024lfm} or even learning the correspondences directly~\cite{teed2020raft,sun2021loftr,tang2022quadtree}. While these bottom-up approaches are capable of achieving pixel-perfect alignment, their reliance on correspondences make them brittle and require salient visual overlap between the images.

With the advent of deep learning, top-down pose estimation models trained on large-scale 3D datasets can learn to estimate relative pose between images with wide baselines~\cite{cai2021extreme,Bezalel2024Extreme}. A key challenge is that the relative pose is often ambiguous. Recent works have explored handling pose estimation probabilistically using factorized distributions~\cite{chen2021wide}, energy-based models~\cite{zhang2022relpose,lin2023relpose++}, or diffusion~\cite{wang2023posediffusion,zhang2024cameras}. More recent approaches have transitioned to distributed ray- or point-based representations of pose to great effect~\cite{wang2023pf,zhang2024cameras,wang2024dust3r}.
Because these methods rely on 3D datasets with limited diversity, finding data for generalization across all scene distributions is an open challenge. 
The current state-of-the art method \duster~ for few-view reconstruction leverages CroCo pre-training \cite{weinzaepfel2023croco} and uses a transformer architecture to predict per-image point maps relative to the camera coordinate frame of the first image. Subsequently, camera poses can be recovered from these predicted point maps.
We view these methods as complementary to our work and in fact, we make direct use of DUSt3R~\cite{wang2024dust3r} as video models can bridge the distribution gap but cannot recover poses by themselves.

\begin{figure}[t]
    \centering
    \includegraphics[width=\linewidth, trim=0 10 0 0, clip]{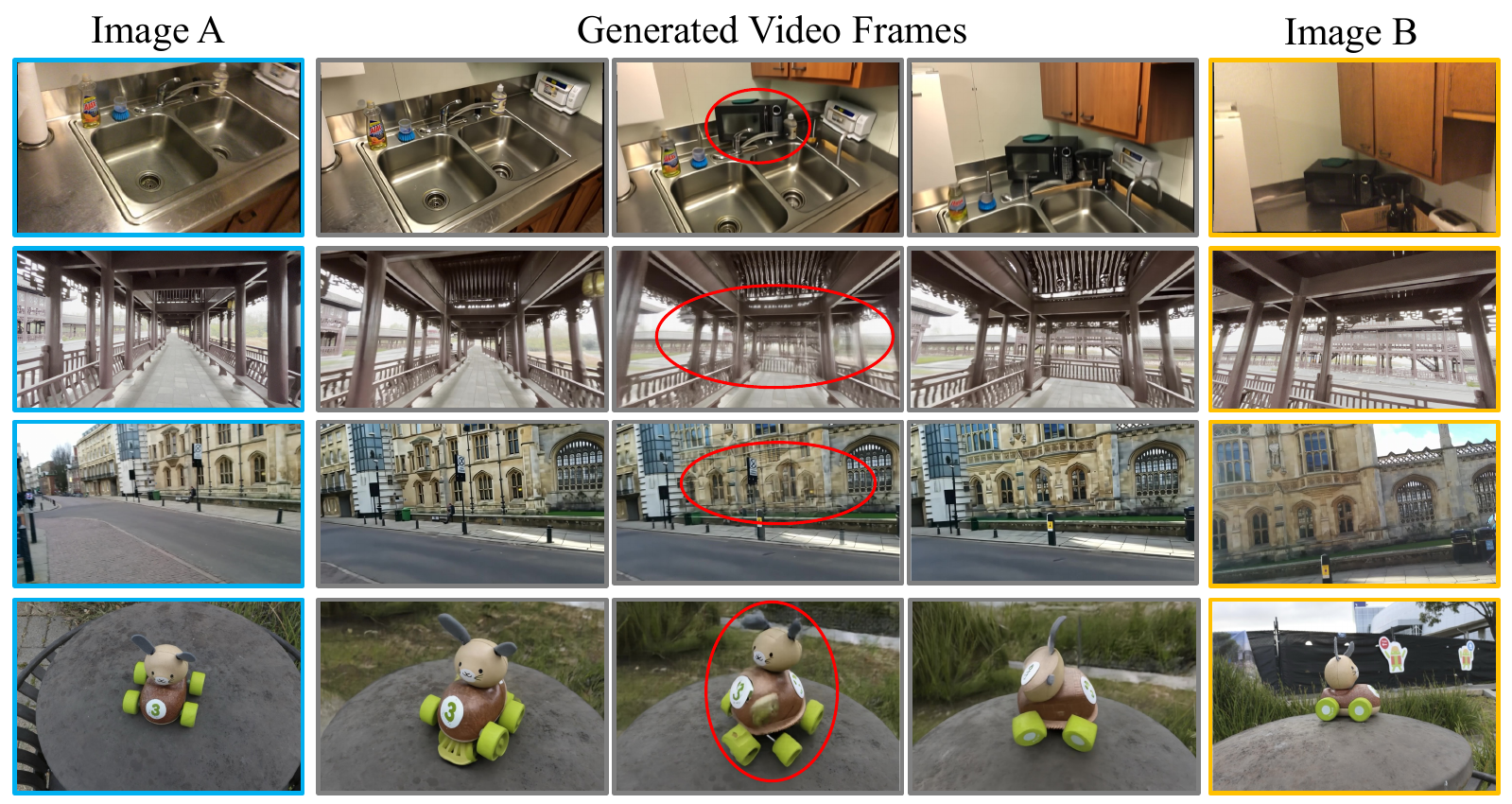}
    \caption{\textbf{Common failure modes of video models.} We show some failure modes of interpolating between two images. In the first row, a microwave suddenly appears over the sink. In the second and third row, the video model morphs and blends images without consistent changes to the underlying scene geometry. In the fourth row, the object's appearance changes in an unrealistic way.
    }
    \label{fig:video_failures}
\end{figure}

\section{Method}
\label{sec:method}

Given two images $I_A$ and $I_B$,  our goal is to recover their relative camera pose. We introduce \method, which leverages off-the-shelf video models to generate the intermediate frames between the two images.
By using these generated frames alongside the original image pair as input to a camera pose estimator, we provide additional context that can improve pose estimation compared to just using the two input images.
A key challenge is that the generated videos 
may contain visual artifacts or implausible motion. Thus, we generate multiple videos which we score using a self-consistency metric to select the best video sample.

\begin{figure*}
    \centering
    \includegraphics[width=\linewidth]{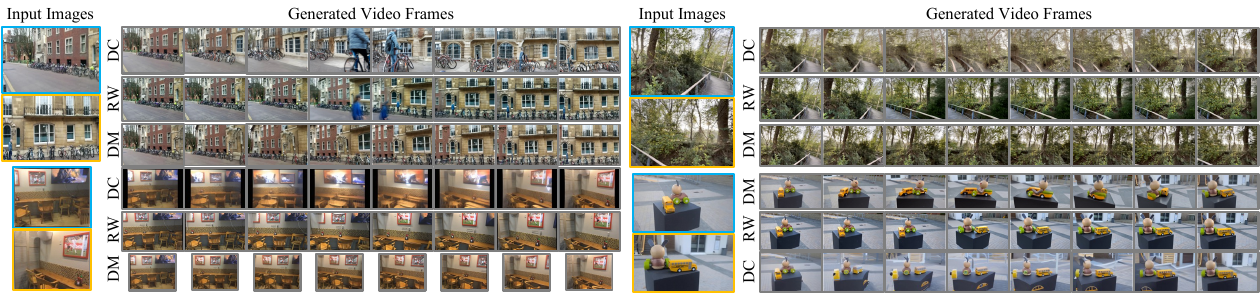}
    \caption{
     Qualitative comparison of the three video models: DynamiCrafter (DC), Runway (RW), and Dream Machine (DM), using the same text prompt for each video model.
    Top left: a pair of images from the Cambridge Landmarks dataset. Prompt: 
    \textcolor{gray}{Dozens of bicycles are parked along the street in front of old brick and stone buildings, with a person walking by and trees in the background.}
    Bottom left: a pair of images from ScanNet. Prompt:
    \textcolor{gray}{A cozy café corner features wooden chairs, framed sports photos, and a TV screen.}
    Top right: a pair from DL3DV-10K. Prompt: 
    \textcolor{gray}{A peaceful morning stroll along a wooden boardwalk surrounded by lush, sunlit greenery.}
    Bottom right: a pair from NAVI. Prompt: 
     \textcolor{gray}{A wooden toy figure with gray ears and green wheels sits next to a small yellow school bus on a black pedestal in an outdoor paved area.}
}
    \label{fig:video_model_comparison}
\end{figure*}

\begin{figure}
    \centering
    \begin{subfigure}[t]{\linewidth}
        \centering
        \includegraphics[width=\linewidth]{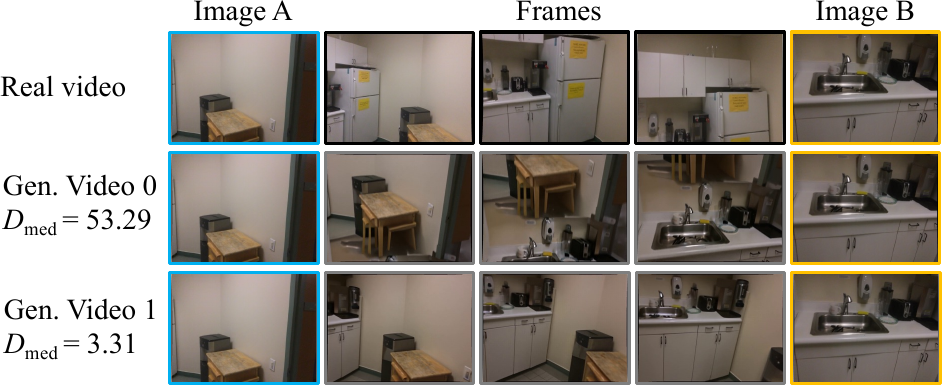}
        \caption{\footnotesize{We take images $A$ and $B$ and generate interpolated videos, (two, Video 0 and Video 1, are shown here for illustration). In this case, the ground truth real video is available, and so we show it at the top for comparison.}}
        \label{fig:distance_fig}
    \end{subfigure}
    \\\vspace{2mm}
    \begin{subfigure}[t]{0.49\linewidth}
        \centering
        \includegraphics[width=\linewidth, trim=0 30 20 10, clip]{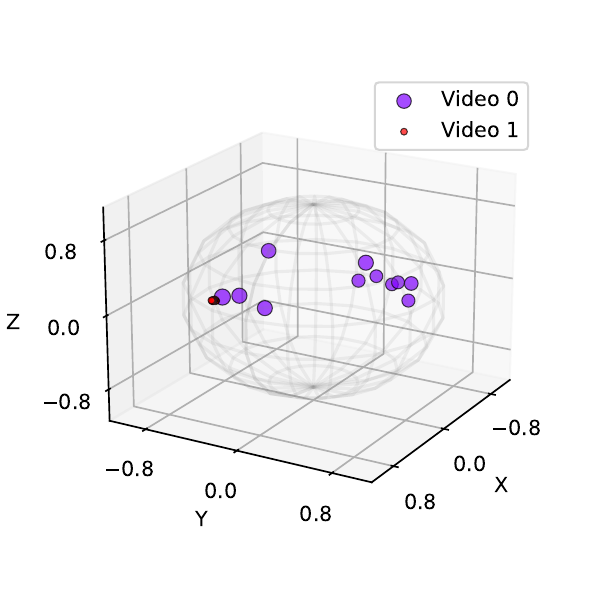}
        \caption{\footnotesize Visualization of predicted rotations using randomly sampled subsets of each generated video on the unit sphere. Note that the samples from Video 1 cluster tightly, and so appear as nearly a single point.}
        \label{fig:rotation_fig}
    \end{subfigure}
    \hfill
    \begin{subfigure}[t]{0.49\linewidth}
        \centering
        \includegraphics[width=\linewidth, trim=0 30 20 10, clip]{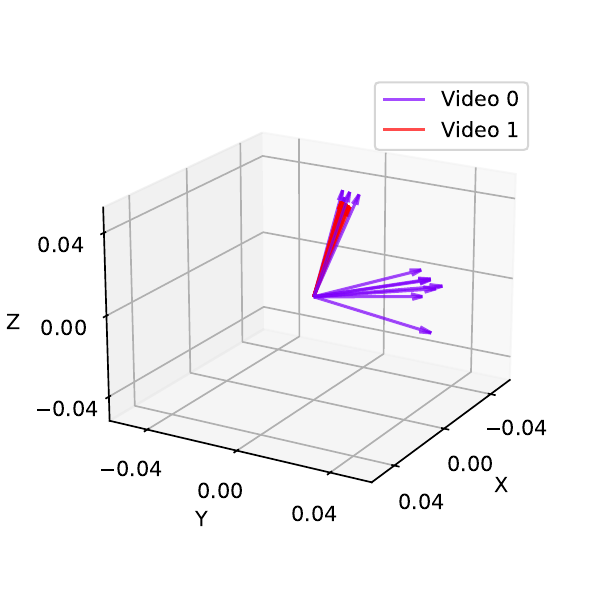}
        \caption{\footnotesize Visualization of predicted translation directions using randomly sampled subsets of frames from Video 0 and Video 1.}
        \label{fig:translation_fig}
    \end{subfigure}
    \caption{\textbf{Self-consistency scores for poses derived from generated videos.} (a) From a pair of input frames $A$ and $B$, we generate several candidate videos from a given video interpolation method. For each video, we sample subsets of frames and compute a relative pose from $A$ to $B$ from each subset ((b) and (c)). We then compute a medoid distance between these samples as a \emph{self-consistency score} for that video, shown to the left of each video in part (a). In this case, Video 0 contains artifacts, and so yields an inconsistent set of poses (and a high medoid distance), which Video 1 is much more natural and produces a more consistent set of poses and a lower medoid distance.
    }
    \label{fig:heuristic}
\end{figure}

\subsection{Preliminaries}
\paragraph{Pose parameterization.}

Given two images \( I_A \) and \( I_B \) associated with ground truth world-to-camera transformations \( T_A \) and \( T_B \):
\begin{equation}
\small
T_A = \begin{bmatrix}
  R_A & t_A \\
  0 & 1
\end{bmatrix}, \quad
T_B = \begin{bmatrix}
  R_B & t_B \\
  0 & 1
\end{bmatrix},
\end{equation}
we aim to recover their relative pose \( T_{\text{rel}} = T_B T_A^{-1} \), where the relative rotation and translation are \( R_{\text{rel}} = R_B R_A^{-1} \) and \( t_{\text{rel}} = t_B - R_{\text{rel}} t_A \), respectively.

The distance between two pose transforms $T_1$ and $T_2$ can be computed by summing their geodesic rotation  and translation angle error. Note that translation angle error makes the distance invariant to scale, and is typically used for pose evaluation.
\begin{equation} 
\small
\text{dist}(T_1, T_2) = \text{dist}_R(R_1, R_2) + \text{dist}_t(R_1, R_2),
\label{eq:geodesic}
\end{equation} 
\begin{equation}
\small
\text{dist}_R(R_1, R_2) = \arccos\left( \dfrac{\text{Trace}(R_2 R_1^\top) - 1}{2} \right), 
\label{eq:rot_error} \end{equation} 
\begin{equation} 
\small
\text{dist}_t(t_1, t_2) = \arccos\left( \left| \dfrac{t_1}{\left\| t_1 \right\|} \cdot \dfrac{t_2}{\left\| t_2 \right\|} \right| \right). \label{eq:trans_error} 
\end{equation}

\paragraph{Camera pose estimator.} 
In the following, we assume a black-box camera pose estimator, that given N images returns estimated relative poses across all N images. In practice, we use \duster~\cite{wang2024dust3r}, but other options could be possible, including non-learning based ones like COLMAP~\cite{schoenberger2016mvs,schoenberger2016sfm}. Although the core \duster~ only reasons about a single image pair, the authors present an extension to compute poses for a set of images based on post-processing optimization over the images' point clouds and poses. In the following, we refer to this extension as \duster. We denote the pose estimator:
\begin{equation}
    f_\text{pose}(\{I_A, I_B, I_{1}, \ldots, I_{N-2}\}) =\hat{T}_\text{B} \hat{T}_\text{A}^{-1} = \hat{T}
\end{equation}
that takes the input pair $I_A$, $I_B$, with optionally additional frames $I_i$, and outputs the relative pose from $I_A$ to $I_B$.

\paragraph{Generative video models.}
We use a generative video model $f_\text{vid}$ capable of interpolating between image frames:
\begin{equation}
\small
    f_\text{vid}(I_A, I_B, p) = [I_1, I_2, \ldots, I_N]
\end{equation}
where $I_1$$=$$I_A$, $I_N$=$I_B$, and $p$ is a text prompt. We consider 3 video models: DynamiCrafter~\cite{xing2025dynamicrafter}, Runway Gen-3 Alpha Turbo~\cite{runway}, and Luma Dream Machine~\cite{lumaDreamMachine}.
We generate multiple samples per input pair $(I_A, I_B)$ by providing different prompts or orderings of the input pair.

\subsection{Self-consistency Score}
\label{sec:score}

Video models generate wildly varying results for similar inputs. This variability is particularly present when doing video interpolation, where a number of camera paths and scene configurations are possible, especially in the low- or no overlap case. Furthermore, the quality of the different samples varies a lot, and artifacts and inconsistencies (e.g., objects appearing/disappearing) are common, as shown in~\cref{fig:video_failures}.
To address these issues, we propose a two-pronged approach:
1) we generate $n$ different videos to account for inherent variability, and 2) we develop a score to identify the video that exhibits the most consistent structure.

\paragraph{Determining consistent videos.} Consider a low quality video that has rapid shot-cuts or inconsistent geometry (\cref{fig:video_failures}). Selecting different subsets of frames from that video would likely produce dramatically different pose estimations. We operationalize this concept by measuring a video's ``self-consistency."

For a given sampled video, we randomly select $m$ sets of $k$ frames (always including the original input images $I_A$ and $I_B$), and calculate the predicted relative pose for each frame subset:
\begin{equation}
    \small
    f_\text{pose}(\{I\}^{(i)}) = \hat{T}^{(i)}.
\end{equation}
We quantify video inconsistency using the medoid distance:
\begin{equation}
    \small
    D_\text{med} = \min_i \frac{1}{m - 1}\sum_{j\not=i} \text{dist}\left(\hat{T}^{(i)}, \hat{T}^{(j)}\right).
\end{equation}
Intuitively, a low medoid distance indicates that every subset of frames produces roughly the same relative pose between $I_A$ and $I_B$, suggesting a consistent video. We illustrate this concept in \cref{fig:heuristic}.

In some degenerate cases, a video that is generated poorly (e.g. only has blurry or uninformative frames) could still have low medoid distance if it consistently makes blatantly incorrect predictions (e.g., always 180 degrees apart). Thus, we found it helpful to bias the metric so that the medoid should not deviate too far from the pose estimated from the original input images alone:
\begin{equation}
    D_\text{total} = D_\text{med} + \text{dist}\left(\hat{T}_\text{med}, f_\text{pose}(\{I_A, I_B\})\right),
\end{equation}
where $\hat{T}_\text{med}$ is the medoid relative pose.

\paragraph{Putting it all together.} We select the video with the lowest $D_\text{total}$, and output as the consensus pose the predicted medoid relative pose $\hat{T}_\text{med}$.

\subsection{Implementation Details}
\begin{figure*}
    \centering
    \includegraphics[width=\linewidth, trim=0 5 0 10, clip]{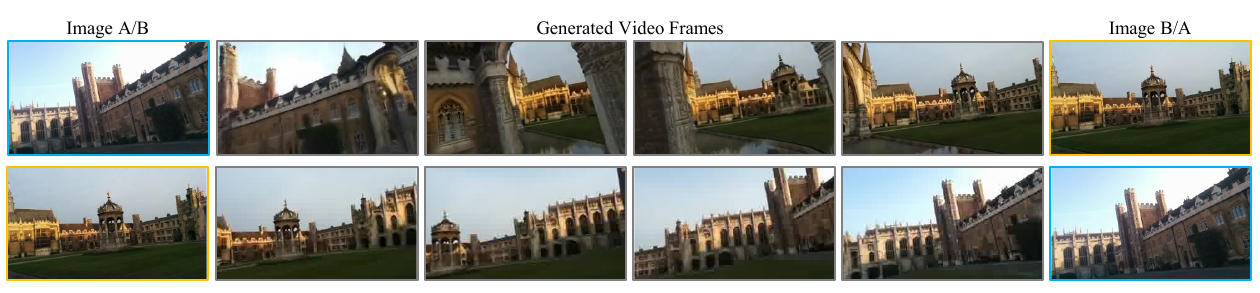}
    \vspace{-5mm}
    \caption{\textbf{Left-to-right bias.} We observed that video models exhibit a tendency to generate similar camera motions (e.g., both left-to-right pans) regardless of the intended direction of interpolation (i.e., transitioning from image A to image B or from image B to image A). This suggests an underlying bias within the model. To mitigate this bias, we swap the order of input images during the generation process. }
    \label{fig:left_right}
\end{figure*}

For each image pair $I_A$ and $I_B$, we use GPT-4o~\cite{achiam2023gpt} to generate two different captions to describe the content of the input image (``Use one sentence to caption these images of the same static scene" and ``Use simple language to specifically include details that describe the same scene shown in these two images in one sentence"). We then use the captions to generate interpolated videos for both the original ($I_A$ to $I_B$) and the flipped order ($I_B$ to $I_A$). We found this flipping to be crucial because video models are often biased toward producing videos that pan to the right as opposed to the left (see \cref{fig:left_right}).

These generated video prompts guide the video models to produce coherent intermediate frames (see \cref{fig:video_model_comparison}). 
Using each of the four generated prompts, we run each video model to interpolate in the specified direction, resulting in a total of $n=4$ generated videos per image pair.
For each generated video, we sample subsets of $k=5$ images (2 original input, 3 generated) to compute candidate poses. In particular, we sample subsets of frames randomly 10 times and once with uniform spacing, for a total of $m=11$ sampled frame subsets per video. For each sample, the $k=5$ frames are provided as input to DUSt3R, and from the resulting poses we compute the medoid as described above.

\begin{figure*}
\begin{center}
\includegraphics[width=\textwidth, trim=0 200 0 0, clip]{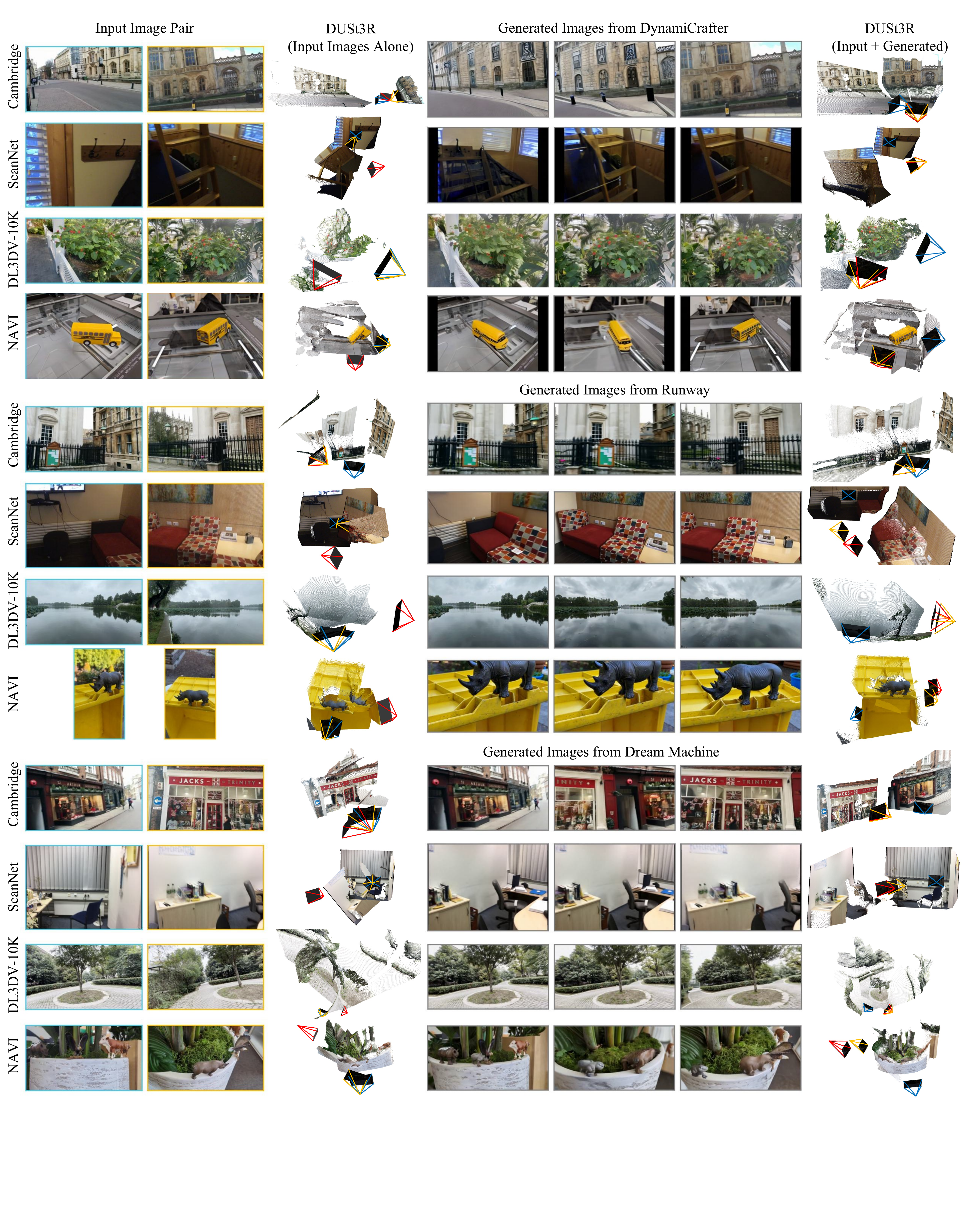}
\end{center}
\vspace{-7pt}
   \caption{ 
\textbf{   Qualitative results of pose estimation from DUSt3R taking only image pair as input and taking additional video frames.
}   We show the input image pair in the first two columns, and the DUSt3R prediction using the image pair alone in the third column. The 3D reconstruction shows the predicted point maps and camera poses for the input images, with the first camera denoted in \textcolor{cyan}{blue}, the second camera in \textcolor{Dandelion}{gold}, and its corresponding ground truth camera in \textcolor{Maroon}{red}, best seen digitally. 
   In columns four to six, we visualize interpolated frames from three different video models. 
   In the last column, we show the DUSt3R pose predictions made using all 5 images, but we are only showing the poses and pointmaps corresponding to the input images for clarity.
   }    
\label{fig:qualitative1}
\end{figure*}

\section{Experiments}

\subsection{Dataset and Benchmark}

We evaluate our method, \method, on challenging inputs from four datasets annotated with ground truth 3D camera poses, covering a diverse range of indoor and outdoor setups.
For each dataset, we selected image pairs by randomly sampling frames within a specified delta yaw range (see below).
This selection ensures challenging pose estimation scenarios with sufficiently large viewpoint changes. Due to the prohibitive cost of running commercial video models, we limit the evaluation to at most 300 image pairs per dataset. We will release the selected indices for reproducibility.

\smallskip
\noindent\textbf{Cambridge Landmarks~\cite{kendall2015posenet}:} This outdoor, scene-scale video dataset captures streets and building facades in Cambridge. 
We utilize a subset of 290 image pairs from \cite{Bezalel2024Extreme} with yaw changes between 50° and 65°.
These pairs feature small to no overlap, with motions characterized predominantly by rotation but minimal camera translation. Thus, we report only rotation metrics for this dataset.

\smallskip
\noindent\textbf{ScanNet~\cite{dai2017scannet}:} An indoor, scene-scale video dataset capturing various indoor environments. We randomly selected 300 image pairs from test 75 scenes, with yaw changes in the range of 50° and 65°. %
    
\smallskip
\noindent\textbf{DL3DV-10K~\cite{ling2024dl3dv}:} A scene-scale, center-facing video dataset comprising over 10,000 videos from 65 types of point-of-interest locations. We randomly selected 300 pairs from 300 outdoor scenes, each with yaw changes ranging from 50° to 90°.
    
\smallskip
\noindent\textbf{NAVI~\cite{jampani2023navi}:} An object-centric, center-facing dataset that includes video and multiview images captured using various camera devices under different environmental conditions. 
We randomly selected 300 pairs from 36 objects, each with yaw changes between 50° and 90°.

While all datasets feature significant viewpoint changes, the center-facing nature of DL3DV-10K and NAVI leads to large overlaps in the view frustrums between input views.
Our experiments indicate that these center-facing datasets are significantly easier for pose prediction than ScanNet and Cambridge Landmarks, which have many outward-facing camera viewpoints.

\begin{table*}[t]
\scriptsize
\centering
\setlength{\tabcolsep}{3pt} %
\begin{tabular}{llccccccccccccccc}
\toprule
                                &                             & \multicolumn{5}{c}{Cambridge Landmarks}                       &  & \multicolumn{9}{c}{ScanNet}                                                                                                                                  \\

                             \cmidrule(lr){3-7} \cmidrule(lr){9-17}
\multirow{2}{*}{Pose estimator} & \multirow{2}{*}{Input data} & \multirow{2}{*}{MRE$\downarrow$} & \multicolumn{3}{c}{R$_\text{acc}\uparrow$} & \multirow{2}{*}{AUC$_{\text{30}}\uparrow$}                        &  & \multirow{2}{*}{MRE$\downarrow$} & \multirow{2}{*}{MTE$\downarrow$} & \multicolumn{3}{c}{R$_\text{acc}\uparrow$}                        & \multicolumn{3}{c}{t$_\text{acc}\uparrow$}                  & \multirow{2}{*}{AUC$_{\text{30}}\uparrow$} \\
\cmidrule(lr){4-6} \cmidrule(lr){11-13} \cmidrule(lr){14-16}
                                &                             &                      & 5°              & 15°             & 30°         &    &  &                      &                      & 5°              & 15°             & 30°            & 5°              & 15°          & 30°          &                         \\
                                \midrule
SIFT+N.N.                     & \multirow{3}{*}{Pair}       & 97.64                & 15.17          & 22.41          & 24.48          & 20.49                   &  & 112.95               & 48.99                & 2.06           & 3.44           & 5.50           & 23.02          & 25.09          & 31.62          & 1.82                    \\
LOFTR                           &                             & 30.30                & 31.38          & 56.55          & 70.00          & 51.63                   &  & 64.46                & 45.49                & 8.33           & 17.00          & 22.00          & 27.00          & 28.33          & 35.33          & 6.43                    \\
DUSt3R                          &                             & 13.28                & 63.45          & 87.24          & 88.97          & 77.23                   &  & 21.31                & 24.72                & 65.33          & 76.33          & 79.00          & 48.33          & 68.33          & 73.67          & 60.34                   \\
 \midrule
 \multirow{3}{*}{Ours (Avg.)}  & DynamiCrafter               & 13.22                & 60.00          & 86.90          & 89.66          & 76.36                   &  & 19.97                & 18.87                & 62.33          & 78.67          & 83.00          & 45.33          & 67.33          & 74.33          & 58.84                   \\
                                & Runway                      & 12.49                & 47.59          & 84.14          & 90.69          & 72.93                   &  & 22.87                & 18.96                & 57.33          & 73.67          & 79.00          & 36.67          & 64.33          & 72.67          & 54.77                   \\
                                & Dream Machine                & 21.85                & 31.38          & 69.66          & 80.00          & 59.39                   &  & 22.44                & 19.82                & 50.33          & 67.00          & 75.00          & 36.67          & 59.33          & 72.33          & 53.00                   \\
\midrule
 \multirow{3}{*}{Ours (Medoid)} & DynamiCrafter               & 12.70                & \textbf{65.17} & 88.97          & 90.34          & 79.00                   &  & 18.96                & 16.42                & 68.00          & \textbf{82.33} & 84.33          & 48.67          & 71.67          & 80.33          & 62.14                   \\
                                & Runway                      & \textbf{10.78}       & 64.83          & \textbf{91.03} & \textbf{94.14} & \textbf{80.59}          &  & 19.93                & 16.31                & 67.67          & 81.33          & 84.33          & \textbf{51.00} & \textbf{72.33} & 80.67          & 61.83                   \\
                                & Dream Machine                & 11.96                & 57.93          & 89.66          & 92.76          & 78.67                   &  & \textbf{17.65}       & \textbf{15.88}       & \textbf{68.67} & 81.33          & \textbf{85.33} & 47.67          & 71.33          & \textbf{82.33} & \textbf{63.06}          \\
\midrule
Oracle & All Video Models & 3.65                 & 90.69          & 96.55          & 98.28          & 92.08                   &  & 5.80                 & 5.00                 & 81.33          & 94.33          & 95.00          & 73.33          & 91.00          & 96.67          & 81.19                  
        
\\ \bottomrule
\end{tabular}
\vspace{-2mm}
\caption{\textbf{Camera pose estimation results on outward-facing datasets (Cambridge Landmarks and ScanNet).} We evaluate the task of pairwise pose estimation. We consider two variants of selection heuristics: averaging poses from randomly sampled frames (Avg.) and selecting the most self-consistent video using our minimal medoid distance metric (Medoid). Our method consistently outperforms DUSt3R on input pairs alone across three video generators. We also present an Oracle baseline that picks the best possible relative pose recovered from all videos generated.}
\label{tab:outward}
\end{table*}

\begin{table*}[t]
\scriptsize
\setlength{\tabcolsep}{3pt} %
\resizebox{\textwidth}{!}{
\begin{tabular}{llcccccccccccccccccc}
\toprule
                                &                             & \multicolumn{9}{c}{DL3DV-10K}                                                                                                                                                & \multicolumn{9}{c}{NAVI}                                                                                                                                     \\
                                \cmidrule(lr){3-11} \cmidrule(lr){12-20}
\multirow{2}{*}{Pose estimator} & \multirow{2}{*}{Input data} & \multirow{2}{*}{MRE$\downarrow$} & \multirow{2}{*}{MTE$\downarrow$} & \multicolumn{3}{c}{R$_\text{acc}\uparrow$}                        & \multicolumn{3}{c}{t$_\text{acc}\uparrow$}                     & \multirow{2}{*}{AUC$_{\text{30°}}$$\uparrow$} &         \multirow{2}{*}{MRE$\downarrow$} & \multirow{2}{*}{MTE$\downarrow$} & \multicolumn{3}{c}{R$_\text{acc}\uparrow$}                        & \multicolumn{3}{c}{t$_\text{acc}\uparrow$}                  & \multirow{2}{*}{AUC$_{\text{30°}}$$\uparrow$} \\ 
\cmidrule(lr){5-7} \cmidrule(lr){8-10} \cmidrule(lr){14-16} \cmidrule(lr){17-19}
                                &                             &                      &                      & 5°              & 15°             & 30°           & 5°              & 15°             & 30°          &                                 &                      &                      & 5°              & 15°             & 30°             & 5°           & 15°          & 30°             &                         \\ \midrule
SIFT+N.N.                     & \multirow{3}{*}{Pair}       & 76.64                & 46.80                & 18.06          & 28.09          & 33.44          & 31.77          & 33.11          & 36.45          & 12.11                   &  107.46               & 45.10                & 4.67           & 6.67           & 7.33           & 16.33          & 17.00          & 19.00          & 3.20                    \\
LOFTR                           &                             & 35.92                & 41.76                & 37.67          & 52.33          & 61.00          & 40.00          & 41.00          & 45.33          & 23.53                   &  71.34                & 51.21                & 6.67           & 14.33          & 19.00          & 24.67          & 25.33          & 29.33          & 4.88                    \\
DUSt3R                          &                             & 10.72                & 13.08                & 39.67          & 87.33          & 94.00          & 55.33          & 83.67          & 89.00          & 66.99                   &  8.65                 & 7.88                 & 68.67          & 92.67          & 94.67          & 69.00          & 92.33          & 95.00          & 78.66                   \\
 \midrule
 \multirow{3}{*}{Ours (Avg.)}  & DynamiCrafter               & 10.45                & 11.30                & 37.33          & 88.67          & 95.00          & 49.33          & 83.67          & 89.67          & 65.76                   &  8.39                 & 7.86                 & 57.00          & 91.33          & 96.00          & 56.00          & 89.00          & 97.00          & 74.33                   \\
                                & Runway                      & 10.27                & 10.86                & 38.67          & 88.67          & 95.33          & 50.33          & 83.00          & 90.33          & 66.32                    & 8.45                 & 8.14                 & 55.33          & 90.00          & 94.67          & 48.33          & 88.00          & 96.33          & 72.79                   \\
                                & Dream Machine                & 10.40                & 11.17                & 35.33          & 86.67          & 94.33          & 46.67          & 83.33          & 89.67          & 64.59                    & 8.58                 & 8.22                 & 55.33          & 91.00          & 95.00          & 56.00          & 89.67          & 95.67          & 74.11                   \\
\midrule
\multirow{3}{*}{Ours (Medoid)}         & DynamiCrafter               & 10.02                & 9.13                 & 38.33          & 87.33          & 95.67          & \textbf{58.33} & \textbf{87.00} & 93.00          & 67.97                   &  8.26                 & 6.57                 & 68.00          & 92.67          & 95.67          & 69.00          & 91.67          & 96.67          & 78.78                   \\
                                & Runway                      & 9.49                 & 8.81                 & \textbf{41.33} & \textbf{90.33} & \textbf{96.67} & 57.33          & 86.67          & 92.33          & \textbf{69.44}          &  8.08                 & \textbf{6.24}        & 67.67          & \textbf{93.67} & \textbf{96.00} & 67.67          & \textbf{93.33} & \textbf{97.00} & 79.02                   \\
                                & Dream Machine                & \textbf{9.13}        & \textbf{8.72}        & \textbf{41.33} & \textbf{90.33} & 96.33          & 57.67          & 86.33          & \textbf{94.67} & 69.11                    & \textbf{7.85}        & 6.51                 & \textbf{69.33} & \textbf{93.67} & 95.33          & \textbf{71.00} & 93.00          & 95.67          & \textbf{79.06}          \\
 \midrule
Oracle & All Video Models                                & 3.99                 & 2.90                 & 71.00          & 98.33          & 100.00         & 86.00          & 97.33          & 98.33          & 85.20                   &  3.05                 & 1.91                 & 90.00          & 98.67          & 99.33          & 94.67          & 98.67          & 99.67          & 91.46                  

\\\bottomrule
\end{tabular}
}
\vspace{-2mm}
\caption{\textbf{Camera pose estimation results on center-facing datasets (DL3DV-10K and NAVI).} DUSt3R exhibits significantly improved performance on these center-facing datasets compared to outward-facing ones. Our method still achieves slightly better results, demonstrating that using a video model does not hinder performance even when DUSt3R is already strong.}
\label{tab:center}
\end{table*}

\subsection{Experimental Variants}

\subsubsection{Baselines}
We compare our method against several pose estimators:

\noindent\textbf{SIFT~\cite{lowe2004distinctive} + Nearest Neighbors:} As a classic geometric baseline, we match SIFT features using nearest neighbors and RANSAC~\cite{fischler1981random} to filter outliers. Using ground truth intrinsics, we compute the essential matrix, from which we extract relative rotations and translations using OpenCV~\cite{opencv_library}.

\smallskip
\noindent\textbf{LOFTR~\cite{sun2021loftr}:} LOFTR uses a transformer to learn semi-dense matches between images. As with the SIFT baseline, we filter outliers and use the correspondences to estimate an essential matrix.

\smallskip
\noindent\textbf{DUSt3R~\cite{wang2024dust3r}:}
DUSt3R is a recent state-of-the-art method for pose estimation and 3D reconstruction from unconstrained image collections. 
Given any number of images as input, DUSt3R reconstructs a dense pointmap for each pair of images. It then jointly optimizes the camera poses and globally aligns the point clouds.

\subsubsection{Variants of our model}

\noindent\textbf{Best Medoid:} 
We use the medoid relative transformation predicted from the generated video with the lowest total medoid distance (see \cref{sec:score}). 

\smallskip
\noindent\textbf{Average:} 
To evaluate the contribution of our self-consistency score using the medoid distance, we also evaluate an approach that takes the average of all $n\cdot m$ predictions from the video model. This tells us whether frames from a video model without any heuristic selection can still help with pose estimation. 

\smallskip
\noindent\textbf{Oracle:} This picks the best possible set of poses with the minimal rotation and translation error among all $n\cdot m$ generated predictions from all three video models. This serves as an upper-bound for a ground-truth heuristic selection.

\subsection{Video Models}

We evaluate three video models (visualized in \cref{fig:video_model_comparison}):

\noindent\textbf{DynamiCrafter~\cite{xing2025dynamicrafter}:} DynamiCrafter is an open-source image animation model enabling video generation and keyframe interpolation. DynamiCrafter is based on a pretrained text-to-video diffusion model and finetuned on WebVid10M \cite{bain2021frozen} for video generation from images and text prompts. %
Given an image pair and text prompt, DynamiCrafter generates 16 frames of resolution $320\times512$.

\smallskip
\noindent\textbf{Runway~\cite{runway}:} Runway Gen-3 Alpha Turbo model is a commercial video generation model to generate video from text and images. The output video has 112  frames of $1280\times768$.

\smallskip
\noindent\textbf{Luma Dream Machine~\cite{lumaDreamMachine}:} Luma Dream Machine is a commercial video generation model that generates video from text and images. The generated video is 114 frames with the same aspect ratio as the input, and approximately one megapixel resolution.

In total, we spent \$5,500 on generating prompts and running the commercial video models.

\subsection{Metrics}

For each pair of images, we evaluate the pose accuracy. We compute the geodesic rotation error and translation angle error using \cref{eq:rot_error,eq:trans_error} respectively. We report the mean rotation error (\textbf{MRE}) and mean translation error (\textbf{MTE}) in degrees. We also evaluate the percentage of rotation (\textbf{R$_\text{acc}$}) and translation (\textbf{t$_\text{acc}$}) errors that are within 5°, 15°, and 30° of the ground truth. Finally, we report the Area-Under-Curve (\textbf{AUC$_{30}$}) from 0° to 30° at 1° thresholds for rotation and translation accuracy following~\cite{jin2021image,wang2023posediffusion}.

\subsection{Quantitative results}
In Table~\ref{tab:outward} and Table~\ref{tab:center}, we present a quantitative evaluation of camera pose estimation on challenging subsets of image pairs on four diverse datasets.

\smallskip
\noindent\textbf{Baseline comparison.} Feature matching-based methods like SIFT+NN and LOFTR struggle when the input pair shares little-to-no overlap as they rely on visual correspondences between overlapping region to estimate camera pose. DUSt3R shows significant improvements over SIFT+NN and LOFTR since it was trained on diverse 3D data without relying solely on explicit feature correspondences.

\smallskip
\noindent\textbf{Performance with Generative Video Models.} 
We find that our method of combining generative video models with DUSt3R consistently enhances performance across all datasets.
Taking the generated frames as additional input to DUSt3R and selecting the most reliable prediction with proposed self-consistency score outperforms only 
relying on the input frame pair alone. This finding holds for 
all three off-the-shelf video models for both rotation and translation.

On outward-facing datasets (Cambridge Landmarks and ScanNet, Table~\ref{tab:outward}), our method significantly reduces pose estimation errors. Notably, on Cambridge Landmarks, mean rotation error decreases from 13.28° to 10.78° using Runway's model, while on ScanNet, mean rotation and translation errors drop from (21.31°,~24.74°) to (17.65°,~15.88°) using Dream Machine.

On the center-facing datasets (DL3DV-10K and NAVI), the improvement is less pronounced but still present, as illustrated in Table~\ref{tab:center}, as these center-facing datasets inherently contain overlapping regions between input views.
On DL3DV-10K dataset, the mean translation error decreased from 13.08° to 8.72° and \text{t$_\text{acc}$}@30° increased from 89\% to 94.67\% using frames from Dream Machine. On the NAVI dataset, the DUSt3R pair only baseline already works well out of the box, but our video model still decreased the mean rotation and translation error by about 1° each.

\smallskip
\noindent\textbf{Effectiveness of self-consistency-aware  score.} 
We observe that simply averaging pose predictions from generated frames leads to worse performance than just taking original image pair as input.
For instance, in Table~\ref{tab:outward} on the Cambridge Landmarks dataset, averaging among the predictions using Dream Machine's frames is even worse than not using a video model at all, with the mean rotation error increasing from $13.28\degree$ to $21.85\degree$.
By using our self-consistency metric, the mean rotation error of predictions with Dream Machine reduces to $11.96\degree$.
This validates the necessity and effectiveness of our medoid-based selection strategy in filtering out low-quality videos and unreliable predictions, thereby preventing degeneration in pose accuracy.

The Oracle outperforms all methods by a wide margin. This implies that with sufficient samples, it is possible for a video generation model to produce frames that are highly informative for pose estimation. It also suggests that there is still significant room for improving the selection method for reliably identifying the best generated frames or videos for pose estimation.

\subsection{Qualitative results}

In \cref{fig:qualitative1}, we visualize qualitative results of using DUSt3R on the input pairs alone compared with using selected generated frames from a video model. We find that all 3 video models are capable of generating informative intermediate images.
We also visualize more video frames from all three video models in \cref{fig:video_model_comparison}.

Please refer to the supplementary materials for more videos, interactive DUSt3R point clouds, and comparisons.

\section{Conclusion}

In this paper, we did a preliminary investigation into how a video model can be used to help pose estimation. 
We developed a heuristic for measuring the self-consistency of a generated video using a medoid-based selection algorithm, and we found that the additional context from the generated videos consistently helped a state-of-the-art pose estimator. 
This finding holds for the 3 recent publicly available video models that we were able to test. There is still significant room for improvement. That our oracle performs so much better than all other approaches reveals that finding a better video selection strategy is a fruitful area of research.
We also found a number of limitations in current-generation  video models. First, they are quite expensive and slow to run, which limited the scope of our investigation. Second, the videos still could not guarantee multi-view consistency. Although our medoid-distance-selection strategy helped alleviate this issue, sometimes all generated videos were low quality. Finally, we found that the video models are quite sensitive to minor changes such as prompts, camera intrinisics, and image aspect ratios.

\paragraph{Acknowledgments} We would like to thank Keunhong Park, Matthew Levine, and Aleksander Hołynski for their feedback and suggestions.

{
    \small
    \bibliographystyle{ieeenat_fullname}
    \bibliography{main}
}

\appendix
\clearpage
\maketitlesupplementary

\section{Qualitative Results}

We provide additional qualitative results, including more examples with videos generated from four different prompts and three video generation models across four datasets. For more visualizations and interactive DUSt3R point clouds, please visit our project page: \url{Inter-Pose.github.io}.

\section{Effectiveness of our method across different yaw changes}

\begin{figure*}
    \centering
    \begin{subfigure}[t]{\linewidth}
        \centering
        \begin{tabular}{ccc}
        $\Delta$ Yaw vs MRE$\downarrow$ & $\Delta$ Yaw vs MTE$\downarrow$ & $\Delta$ Yaw vs AUC$_{30\degree}\uparrow$ \\
        \includegraphics[width=0.3\linewidth, trim=0 0 0 0, clip]{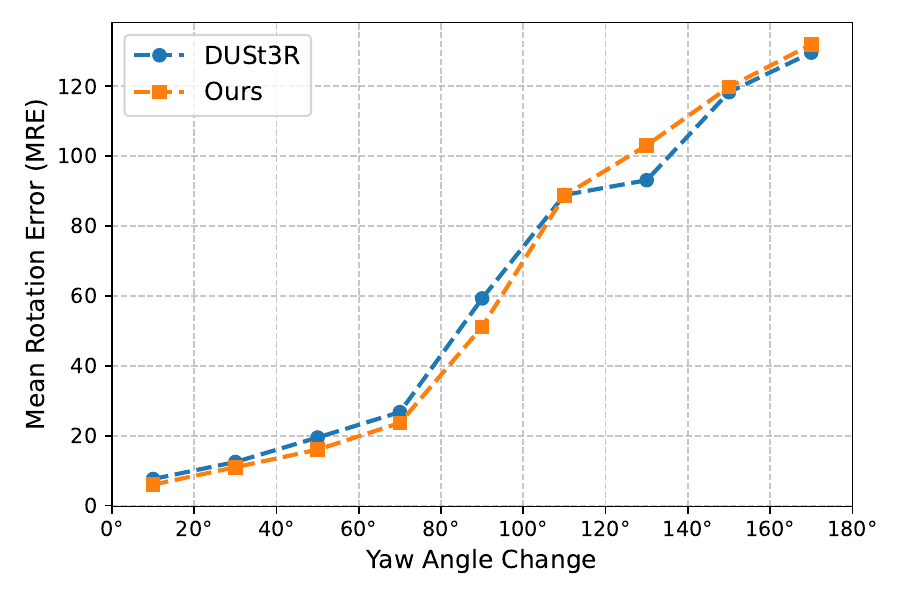} &
        \includegraphics[width=0.3\linewidth, trim=0 0 0 0, clip]{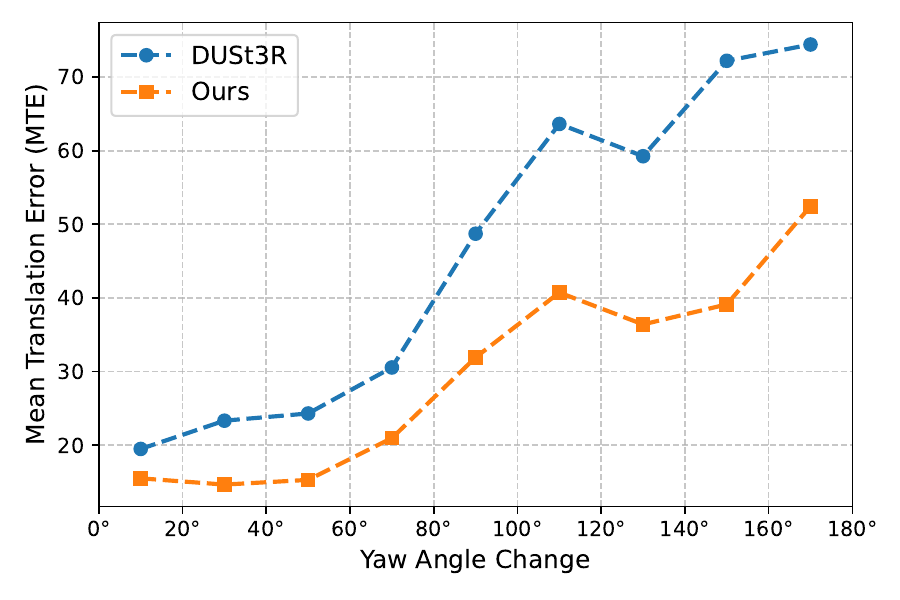} &
        \includegraphics[width=0.3\linewidth, trim=0 0 0 0, clip]{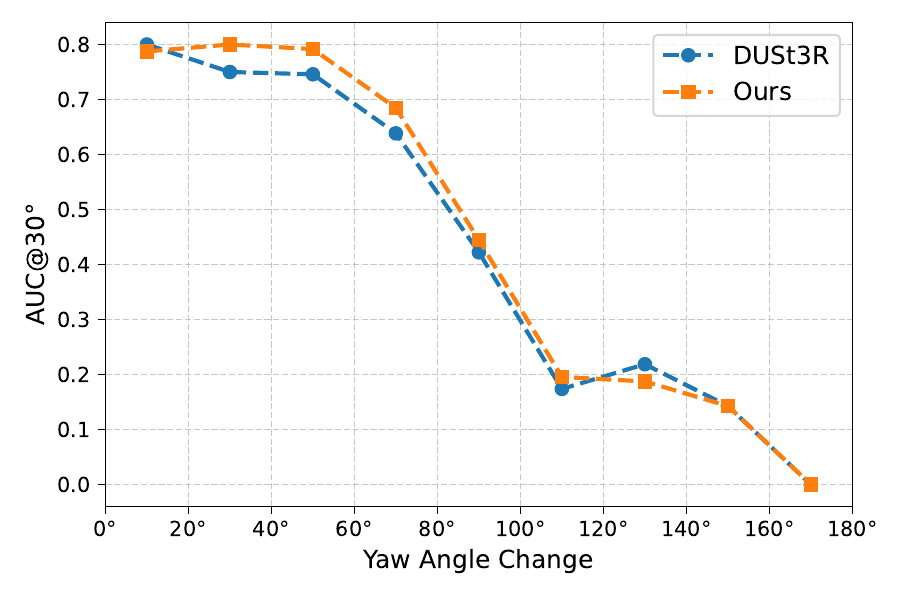}
        \end{tabular}
        \caption{
        Camera Pose Estimation Performance vs. Yaw Angle Change on the ScanNet Dataset.
        }
    \end{subfigure}
    \\\vspace{2mm}
    \begin{subfigure}[t]{\linewidth}
        \centering
        \begin{tabular}{ccc}
        $\Delta$ Yaw vs MRE$\downarrow$ & $\Delta$ Yaw vs MTE$\downarrow$ & $\Delta$ Yaw vs AUC$_{30\degree}\uparrow$ \\
        \includegraphics[width=0.3\linewidth, trim=0 0 0 0, clip]{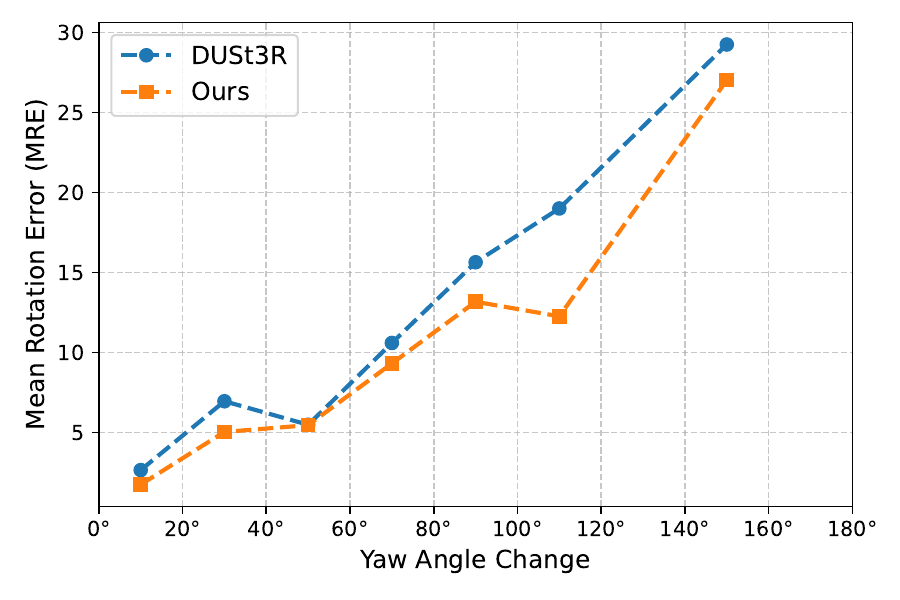}&
        \includegraphics[width=0.3\linewidth, trim=0 0 0 0, clip]{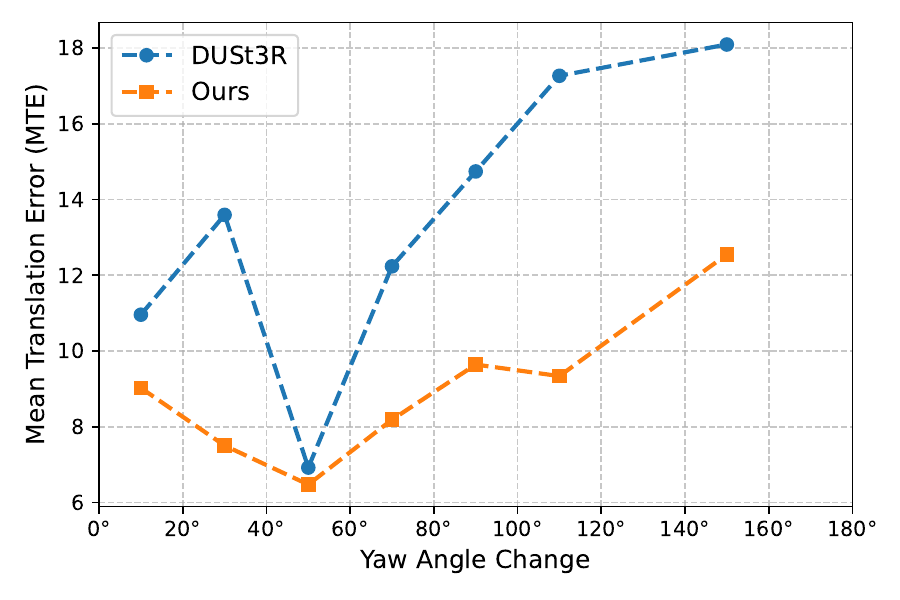}&
        \includegraphics[width=0.3\linewidth, trim=0 0 0 0, clip]{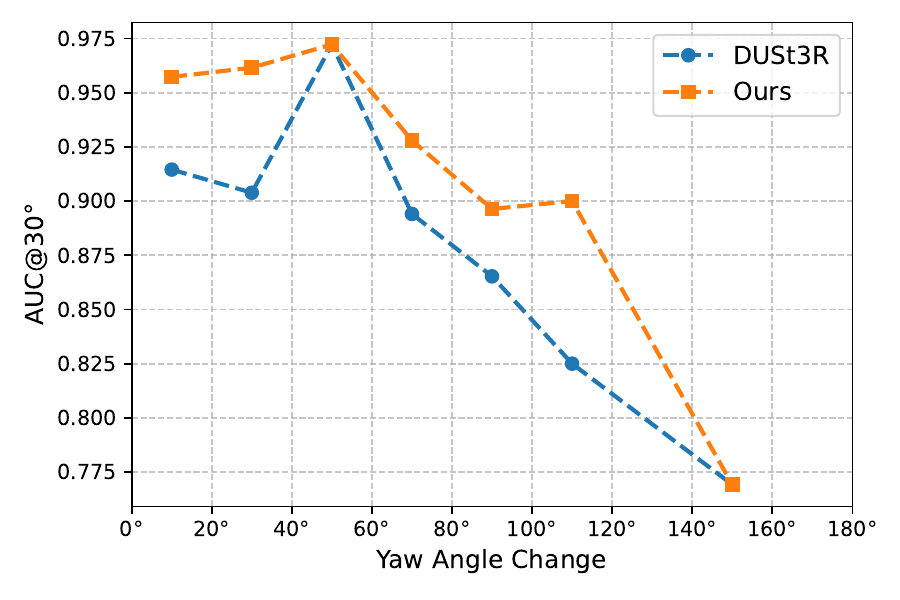}
        \end{tabular}
        \caption{
        Camera Pose Estimation Performance vs. Yaw Angle Change on the DL3DV-10K Dataset.
        }
    \end{subfigure}
    \caption{
    \textbf{Camera Pose Estimation Performance vs. Yaw Angle Change on the ScanNet and DL3DV-10K Datasets.}
    Comparison of Mean Rotation Error (MRE), Mean Translation Error (MTE), and Area Under Curve at 30° (AUC$_{30\degree}$) across different yaw angle change intervals (0°, 20°, 40°, 60°, etc.) 
    Each data point represents the average value of the respective metric within a specific yaw angle range.
    Our method consistently achieves lower errors than DUSt3R for yaw angle changes below $110\degree$ on both datasets.
    Due to the limited number of sample pairs with yaw angle changes larger than $120\degree$ in the DL3DV-10K dataset, we report the results averaged over the [120°, 180°] range.
    }
    \label{fig:yaw}
\end{figure*}

In addition to the small overlapping pairs with yaw changes in the ranges of $[50\degree, 65\degree]$ for outward-facing datasets and $[50\degree, 90\degree]$ for center-facing datasets, as described in the main paper, we conducted further experiments to evaluate the effectiveness of our proposed method on image pairs with either significant overlap or no overlap. These experiments specifically examine the impact of varying yaw angle changes between image pairs.

\smallskip
\noindent\textbf{ScanNet~\cite{dai2017scannet}:}
For this outward-facing, indoor dataset, we sampled 200 pairs with yaw changes in the range of $[0\degree, 50\degree]$ to represent pairs with large overlap, and 200 pairs with yaw changes in the range of $[65\degree, 180\degree]$ to represent non-overlapping pairs.

\smallskip
\noindent\textbf{DL3DV-10K~\cite{ling2024dl3dv}:}
This is a dataset consisting of outdoor scenes with center-facing camera viewpoints. We sampled 200 large-overlap pairs (with yaw changes in the range $[0\degree, 50\degree]$ and 200 pairs with larger yaw changes in the range $[90\degree, 180\degree]$.

For each pair, we the settings described in the main paper by generating four videos using Dream Machine. 
For each video, We randomly selected 11 subsets of 3 frames, along with the original image pair, and used these subsets as input to the DUSt3R pose estimator.
We then computed the total medoid distance of the predicted relative transformations and selected the prediction with the lowest distance as the final relative pose estimate.

In \cref{fig:yaw}, we present camera pose estimation performance vs.\ yaw angle change using the metrics of mean rotation error (MRE), mean translation error (MTE), and AUC$_{30\degree}$. 
As the yaw angle between input image pairs increases, the overlap between images decreases, resulting in higher MRE and MTE for both DUSt3R and our method.
our method consistently achieves lower errors than DUSt3R for yaw changes below $110\degree$ on both the ScanNet and DL3DV-10K datasets.

We provide quantitative results with more metrics on ScanNet in Tables~\ref{tab:supp_scannet_yaw1} and~\ref{tab:supp_scannet_yaw2}.
For large-overlap pairs, our method, which incorporates generated frames from the video model, outperforms DUSt3R (when DUSt3R only uses the input image pair). 
Specifically, the mean rotation and translation errors decreased from (11.33°,~22.50°) to (9.12°,~15.75°) when using Dream Machine. 
For non-overlapping pairs, adding the generated video as input to the pose estimator yields comparable performance to using only the original image pair. This may be due to the ambiguity and multiple possibilities inherent in pairs with no overlap.

Quantitative results for DL3DV-10K are shown in Tables~\ref{tab:supp_scannet_dl3dv_yaw1} and~\ref{tab:supp_scannet_dl3dv_yaw2}.
For large-overlap pairs, our method (using the generated frames from generative videos obtains better results than DUSt3R, reducing mean rotation and translation errors from (4.28°,~11.04°) to (3.23°,~8.16°).
For pairs with yaw changes in $[90\degree, 180\degree]$, the center-facing nature of the DL3DV-10K dataset still results in some overlapping regions. Incorporating the generated video as input improves performance by increasing $\text{R}_{\text{acc}}@30\degree$ and $\text{T}_{\text{acc}}@30\degree$ from (85.50\%,~87.00\%) to (89.50\%,~91.50\%). 
These results also indicate that center-facing datasets like DL3DV-10K are significantly easier for pose prediction than ScanNet and Cambridge Landmarks, which have many outward-facing camera viewpoints.

\section{Results with MASt3R}

MASt3R~\cite{leroy2025grounding}, a recent follow-up method to DUSt3R, follows a similar backbone and training scheme as DUSt3R but incorporates additional heads to produce local features and facilitate feature matching. With these enhancements, MASt3R can produce more accurate pose estimates compared to DUSt3R, particularly when the input pair exhibits overlap and sufficient correspondences are available.

In Table~\ref{tab:mast3r-summary}, we shows the results for MASt3R using the original image pair, as well as using our method (based on the MASt3R pose estimator) which uses generated frames as input to MASt3R and selects the most reliable prediction based on the medoid distance metric.
Comprehensive results with more evaluation metrics can be found in Tables~\ref{tab:mast3r_outward} and~\ref{tab:mast3r_inward}.

\begin{figure*}[ht!]
    \centering

    \begin{tabular}{ccc}
        \textbf{Input pair} & \textbf{DUSt3R} & \textbf{MASt3R} \\
        \begin{subfigure}[t]{0.33\textwidth}
        \centering
        \includegraphics[width=\linewidth]{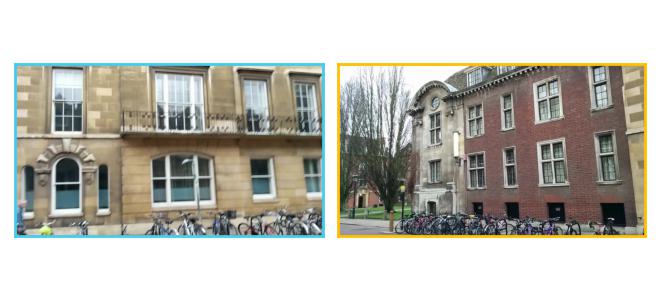}
        \end{subfigure}
        &
        \begin{subfigure}[t]{0.33\textwidth}
            \centering
            \includegraphics[width=0.8\linewidth]{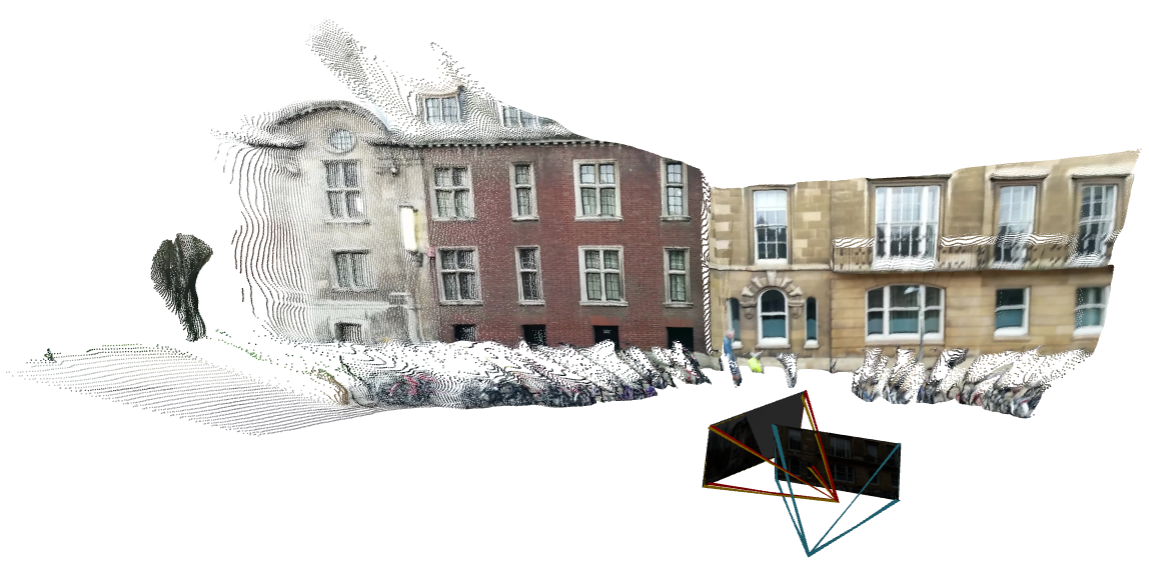}
        \end{subfigure}
        &
        \begin{subfigure}[t]{0.33\textwidth}
            \centering
            \includegraphics[width=0.6\linewidth]{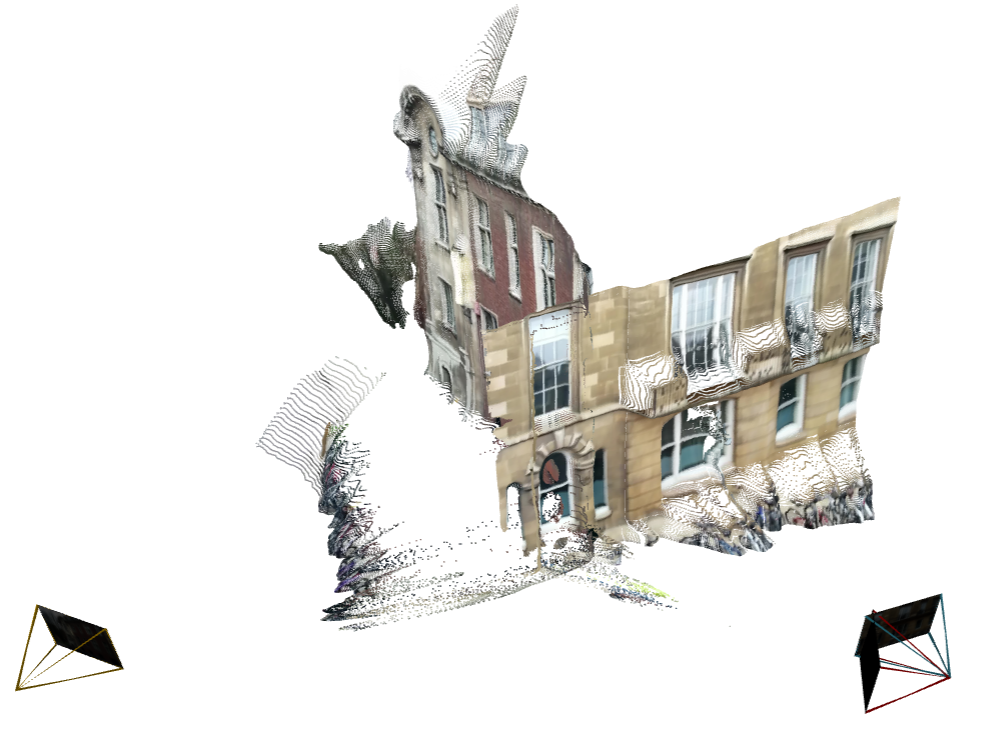}
        \end{subfigure}
        \\
        \begin{subfigure}[t]{0.33\textwidth}
        \centering
        \includegraphics[width=\linewidth]{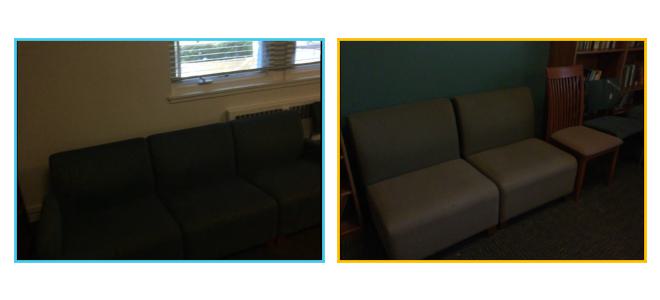}
        \end{subfigure}
        &
        \begin{subfigure}[t]{0.33\textwidth}
            \centering
            \includegraphics[width=0.8\linewidth]{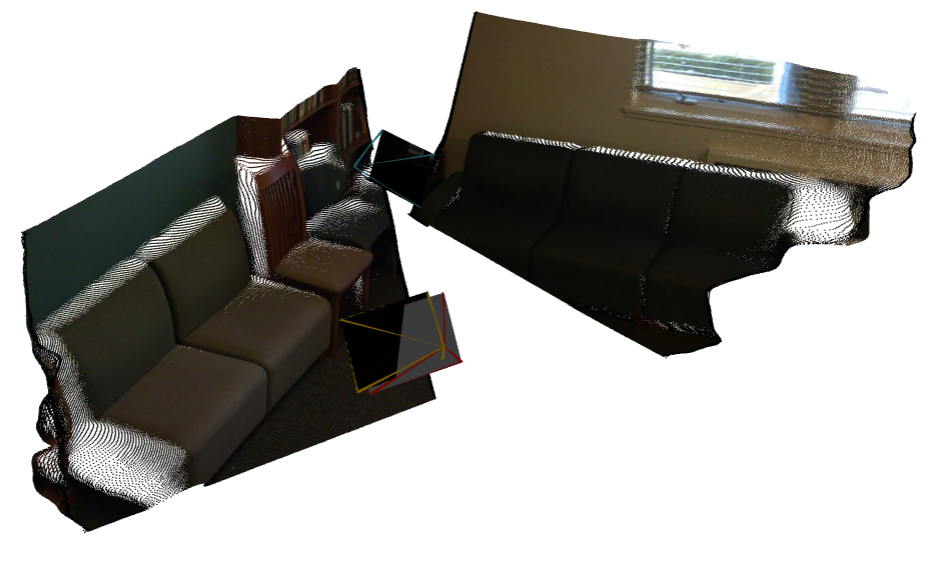}
        \end{subfigure}
        &
        \begin{subfigure}[t]{0.33\textwidth}
            \centering
            \includegraphics[width=0.5\linewidth]{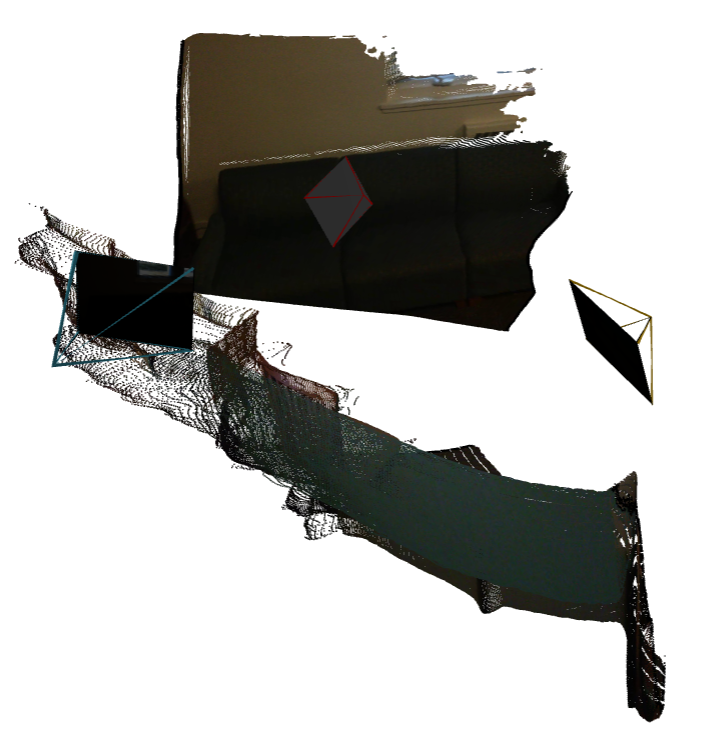}
        \end{subfigure}
    \end{tabular}

    \caption{\textbf{Failure examples of MASt3R.} We show instances where MASt3R fails to accurately predict poses on non-overlapping pairs from the Cambridge Landmarks (top row) and ScanNet (bottom row) datasets. 
    MASt3R relies on feature matching for pose refinement, which is insufficient and less reliable when pairs lack overlapping regions.
    In contrast, DUSt3R demonstrates greater robustness in these scenarios.}
    \label{fig:mast3r_fail}
\end{figure*}

On the Cambridge Landmarks and ScanNet datasets, many image pairs feature outward-facing camera viewpoints and have no overlap. This lack of overlap and correspondence results in MASt3R exhibiting performance that is significantly worse than that of DUSt3R, especially on the Cambridge Landmarks dataset. 
As shown in Figure~\ref{fig:mast3r_fail}, MASt3R completely fails in scenarios with no overlap.
Our method, with MASt3R as the pose estimator, still achieves improvements on both outward-facing datasets. Specifically, it significantly reduces the mean rotation error from $36.55\degree$ to $27.47\degree$ on the Cambridge Landmarks dataset and increases the AUC at $30\degree$ from $55.10\%$ to $58.28\%$ on ScanNet dataset when using video frames generated by Dream Machine.

On the DL3DV-10K and NAVI datasets, which are center-facing datasets where image pairs always share overlapping regions even with large camera viewpoint changes, MASt3R performs significantly better than DUSt3R. Reliable matches can be found in these pairs due to the overlapping regions sampled from center-facing datasets. Given the almost perfect performance of MASt3R on these datasets, our method, which takes video frames as additional input, achieves comparable results to MASt3R when using only image pairs on DL3DV-10K. Additionally, it obtains slight improvements on the NAVI dataset by decreasing the mean rotation and translation errors from (5.59°,~5.23°) to (5.28°,~5.20°) when using generated videos from Runway.

\begin{table*}[tb]
\centering
\resizebox{\textwidth}{!}{
\begin{tabular}{llcclccclccclccc}
\toprule
                               &                  & \multicolumn{6}{c}{Outward-facing datasets}                                                                                        &  & \multicolumn{7}{c}{Center-facing datasets}                                                                                                               \\
                               \cmidrule(lr){3-8} \cmidrule(lr){10-16}
                               &                  & \multicolumn{2}{c}{Cambridge}           &  & \multicolumn{3}{c}{ScanNet}                                                 &  & \multicolumn{3}{c}{DL3DV-10K}                                         &  & \multicolumn{3}{c}{Navi}                                         \\ \cmidrule(lr){3-4} \cmidrule(lr){6-8} \cmidrule(lr){10-12} \cmidrule(lr){14-16}
Pose estimator                 & Input data       & MRE$\downarrow$                     & AUC$_{\text{30°}}$$\uparrow$                  &  & MRE$\downarrow$                     & MTE$\downarrow$                     & AUC$_{\text{30°}}$$\uparrow$                  &  & MRE$\downarrow$                    & MTE$\downarrow$                    & AUC$_{\text{30°}}$$\uparrow$                  &  & MRE$\downarrow$                    & MTE$\downarrow$                    & AUC$_{\text{30°}}$$\uparrow$                  \\ \midrule
Dust3r                         & Pair             & 13.28                   & 77.23                   &  & 21.31                   & 24.72                   & 60.34                   &  & 10.72                  & 13.08                  & 66.99                   &  & 8.65                   & 7.88                   & 78.66                   \\
\multirow{3}{*}{Ours (DUSt3R)} & DynamiCrafter    & 12.70                   & \cellcolor{orange}79.00 &  & \cellcolor{orange}18.96 & 16.42                   & \cellcolor{orange}62.14 &  & 10.02                  & 9.13                   & 67.97                   &  & 8.26                   & 6.57                   & 78.78                   \\
                               & Runway           & \cellcolor{red}10.78    & \cellcolor{red}80.59    &  & 19.93                   & \cellcolor{orange}16.31 & 61.83                   &  & \cellcolor{orange}9.49 & \cellcolor{orange}8.81 & \cellcolor{red}69.44    &  & \cellcolor{orange}8.08 & \cellcolor{red}6.24    & \cellcolor{orange}79.02 \\
                               & Dream Machine    & \cellcolor{orange}11.96 & 78.67                   &  & \cellcolor{red}17.65    & \cellcolor{red}15.88    & \cellcolor{red}63.06    &  & \cellcolor{red}9.13    & \cellcolor{red}8.72    & \cellcolor{orange}69.11 &  & \cellcolor{red}7.85    & \cellcolor{orange}6.51 & \cellcolor{red}79.06    \\
Oracle                         & All Video Models & 3.65                    & 92.08                   &  & 5.80                    & 5.00                    & 81.19                   &  & 1.35                   & 1.05                   & 95.83                   &  & 2.23                   & 1.67                   & 92.90                   \\ \midrule
Mast3r                         & Pair             & 36.55                   & 55.69                   &  & 24.35                   & 17.93                   & 55.10                   &  & \cellcolor{red}4.13    & \cellcolor{red}3.88    & \cellcolor{red}87.22    &  & 5.59                   & \cellcolor{orange}5.23 & 80.84                   \\
\multirow{3}{*}{Ours (MASt3R)} & DynamiCrafter    & 31.43                   & 60.03                   &  & 21.97                   & 16.48                   & \cellcolor{orange}57.90 &  & 4.49                   & 4.04                   & 85.86                   &  & \cellcolor{orange}5.29 & 5.61                   & 80.21                   \\
                               & Runway           & \cellcolor{orange}29.04 & \cellcolor{red}63.57    &  & \cellcolor{orange}21.68 & \cellcolor{orange}15.28 & 57.19                   &  & \cellcolor{orange}4.17 & \cellcolor{orange}4.01 & \cellcolor{orange}86.79 &  & \cellcolor{red}5.28    & \cellcolor{red}5.20    & \cellcolor{red}81.63    \\
                               & Dream Machine    & \cellcolor{red}27.47    & \cellcolor{orange}63.14 &  & \cellcolor{red}19.91    & \cellcolor{red}15.05    & \cellcolor{red}58.28    &  & 4.30                   & 4.21                   & 85.88                   &  & 5.66                   & 5.45                   & \cellcolor{orange}81.42 \\
Oracle                         & All Video Models & 3.65                    & 92.08                   &  & 5.80                    & 5.00                    & 81.19                   &  & 1.35                   & 1.05                   & 95.83                   &  & 2.23                   & 1.67                   & 92.90                  
\\ \bottomrule
\end{tabular}
}
\caption{\textbf{Camera pose estimation results on outward-facing datasets (Cambridge and ScanNet) and center-facing datasets (DL3DV-10K and NAVI).} 
We evaluate our method based on two pose estimators DUSt3R and MASt3R. 
MASt3R demonstrates significantly improved performance on these center-facing datasets compared to outward-facing ones. 
Our method consistently outperforms both DUSt3R and MASt3R on outward-facing datasets, and obtains comparable results on center-facing datasets, demonstrating that using a video model does not hinder performance even when DUSt3R and MASt3R are already strong. }
\label{tab:mast3r-summary}
\end{table*}

\section{Ablation Study}

\begin{table*}[]
\resizebox{\textwidth}{!}{
\begin{tabular}{llcccclccclccclccc}
\toprule
\multirow{2}{*}{Pose estimator} & \multirow{2}{*}{Input data} & \multicolumn{2}{c}{Distance metric}         & \multicolumn{2}{c}{Cambridge}                       &  & \multicolumn{3}{c}{ScanNet}                                                   &  & \multicolumn{3}{c}{DL3DV-10K}                                                &  & \multicolumn{3}{c}{NAVI}                                                    \\
\cmidrule(lr){3-4} \cmidrule(lr){5-6}  \cmidrule(lr){8-10} \cmidrule(lr){12-14}  \cmidrule(lr){16-18}
                                &                             & $D_{\text{med}}$               & $D_{\text{bias}}$            & MRE$\downarrow$         & AUC$_{\text{30}}\uparrow$ &  & MRE$\downarrow$         & MTE$\downarrow$         & AUC$_{\text{30}}\uparrow$ &  & MRE$\downarrow$         & MTE$\downarrow$        & AUC$_{\text{30}}\uparrow$ &  & MRE$\downarrow$        & MTE$\downarrow$        & AUC$_{\text{30}}\uparrow$ \\ \midrule
DUSt3R                          & Pair                        & --                   & --                   & 13.28                   & 77.23                     &  & 21.31                   & 24.72                   & 60.34                     &  & 10.72                   & 13.08                  & 66.99                     &  & 8.65                   & 7.88                   & \cellcolor{orange}78.66   \\
\multirow{3}{*}{Ours}           & DynamiCrafter               & $\checkmark$                & $\checkmark$                & \cellcolor{orange}12.70  & \cellcolor{orange}79.00      &  & \cellcolor{orange}18.96 & \cellcolor{red}16.42    & \cellcolor{orange}62.14   &  & \cellcolor{orange}10.02 & \cellcolor{orange}9.13 & \cellcolor{orange}67.97   &  & \cellcolor{orange}8.26 & \cellcolor{orange}6.57 & \cellcolor{red}78.78      \\
                                & DynamiCrafter               & $\checkmark$                & \multicolumn{1}{l}{} & \cellcolor{red}12.58    & \cellcolor{red}80.31      &  & \cellcolor{red}18.26    & \cellcolor{orange}16.58 & \cellcolor{red}62.94      &  & \cellcolor{red}9.42     & \cellcolor{red}8.93    & \cellcolor{red}68.89      &  & \cellcolor{red}7.12    & \cellcolor{red}6.31    & 78.23                     \\
                                & DynamiCrafter               & \multicolumn{1}{l}{} & $\checkmark$                & 12.88                   & 77.32                     &  & 20.25                   & 21.91                   & 60.43                     &  & 10.48                   & 12.34                  & 67.24                     &  & 8.78                   & 8.05                   & 78.00                     \\  \midrule
DUSt3R                          & Pair                        & --                   & --                   & 13.28                   & 77.23                     &  & 21.31                   & 24.72                   & 60.34                     &  & 10.72                   & 13.08                  & 66.99                     &  & 8.65                   & 7.88                   & \cellcolor{orange}78.66   \\
\multirow{3}{*}{Ours}           & Runway                      & $\checkmark$                & $\checkmark$                & \cellcolor{orange}10.78 & \cellcolor{orange}80.59   &  & \cellcolor{red}19.93    & \cellcolor{red}16.31    & \cellcolor{red}61.83      &  & \cellcolor{orange}9.49  & \cellcolor{red}8.81    & \cellcolor{orange}69.44   &  & \cellcolor{orange}8.08 & \cellcolor{orange}6.24 & \cellcolor{red}79.02      \\
                                & Runway                      & $\checkmark$                & \multicolumn{1}{l}{} & \cellcolor{red}10.77    & \cellcolor{red}80.91      &  & 21.27                   & \cellcolor{orange}16.66 & \cellcolor{orange}61.33   &  & \cellcolor{red}9.11     & \cellcolor{orange}9.26 & \cellcolor{red}69.87      &  & \cellcolor{red}6.70     & \cellcolor{red}6.15    & 78.36                     \\
                                & Runway                      & \multicolumn{1}{l}{} & $\checkmark$                & 12.13                   & 78.52                     &  & \cellcolor{orange}20.68 & 18.78                   & 61.29                     &  & 10.08                   & 12.13                  & 68.04                     &  & 8.18                   & 7.37                   & \cellcolor{orange}78.66   \\  \midrule
DUSt3R                          & Pair                        & --                   & --                   & 13.28                   & 77.23                     &  & 21.31                   & 24.72                   & 60.34                     &  & 10.72                   & 13.08                  & 66.99                     &  & 8.65                   & 7.88                   & 78.66                     \\
\multirow{3}{*}{Ours}           & Dream Machine               & $\checkmark$                & $\checkmark$                & \cellcolor{orange}11.96 & \cellcolor{orange}78.67   &  & \cellcolor{red}17.65    & \cellcolor{orange}15.88 & \cellcolor{red}63.06      &  & \cellcolor{orange}9.13  & \cellcolor{orange}8.72 & \cellcolor{orange}69.11   &  & \cellcolor{orange}7.85 & \cellcolor{red}6.51    & \cellcolor{red}79.06      \\
                                & Dream Machine               & $\checkmark$                & \multicolumn{1}{l}{} & 19.37                   & 71.63                     &  & \cellcolor{orange}18.28 & \cellcolor{red}15.49    & \cellcolor{orange}62.89   &  & \cellcolor{red}8.60      & \cellcolor{red}8.48    & \cellcolor{red}70.10       &  & \cellcolor{red}7.53    & \cellcolor{orange}6.66 & 78.34                     \\
                                & Dream Machine               & \multicolumn{1}{l}{} & $\checkmark$                & \cellcolor{red}11.25    & \cellcolor{red}79.08      &  & 20.24                   & 19.45                   & 60.80                     &  & 10.15                   & 11.98                  & 67.70                     &  & 8.36                   & 7.59                   & \cellcolor{orange}78.70   
\\ \bottomrule
\end{tabular}
}
\caption{\textbf{Abltion study of distance metrics.} 
Our proposed distance metric incorporates both the medoid distance $D_{\text{med}}$ and the bias distance $D_{\text{bias}}$, where $D_{\text{bias}}$ is defined as 
$D_{\text{bias}} = \text{dist}\left(\hat{T}_\text{med}, f_\text{pose}(\{I_A, I_B\})\right)$.
We perform an ablation study to evaluate the contribution of each distance term. While $D_{\text{med}}$ and the total distance yield comparable results across most datasets and video models, solely considering $D_{\text{med}}$ leads to significantly worse performance on the Cambridge dataset when using the Dream Machine video model. 
Incorporating the total distance enhances generalization ability and robustness across various datasets and video models.
}
\label{tab:ablation_distance}
\end{table*}

\begin{table*}[tb]
\centering
\resizebox{\textwidth}{!}{
\begin{tabular}{llllccccccccc}
\toprule
\multirow{2}{*}{Pose estimator} & \multirow{2}{*}{Input data}    & \multirow{2}{*}{\# Images} & \multirow{2}{*}{\# Samples} & \multirow{2}{*}{MRE$\downarrow$} & \multirow{2}{*}{MTE$\downarrow$} & \multicolumn{3}{c}{R$_\text{acc}\uparrow$}                        & \multicolumn{3}{c}{t$_\text{acc}\uparrow$}                  & \multirow{2}{*}{AUC$_{\text{30}}\uparrow$} \\
\cmidrule(lr){7-9}\cmidrule(lr){10-12}
                                &                                &                                  & &                     &                      & 5°              & 15°             & 30°            & 5°              & 15°          & 30°               \\ \midrule
DUSt3R                          & Pair                           & 2+0 & 1                                & 21.31                & 24.72                & 65.33                & 76.33                & 79.00                & 48.33                & 68.33                & 73.67                & 60.34                   \\ 
\midrule
\multirow{5}{*}{Ours}    & \multirow{5}{*}{Pair+Dream Machine} & 2+1   & 11                             & 20.41                & 16.93                & 67.00          & 79.00          & 81.67          & 50.00          & \textbf{71.33} & 81.33          & 62.08                   \\
                                &                                & 2+3  & 11                  & \textbf{17.65}       & \textbf{15.88}       & \textbf{68.67} & 81.33          & \textbf{85.33} & 47.67          & \textbf{71.33} & \textbf{82.33} & \textbf{63.06}          \\
                                &                                & 2+8  & 11                             & 17.98                & 16.39                & 66.00          & 81.33          & 85.00          & \textbf{50.67} & 70.67          & 81.33          & 61.96                   \\
                                &                                & 2+38  & 11                             & 18.43                & 16.70                 & 65.00             & \textbf{82.33} & \textbf{85.33} & 50.33          & \textbf{71.33} & 79.33          & 62.76                   \\
                                &                                & 2+114 & 1                            & 17.77                & 17.05                & 65.67          & \textbf{82.33} & \textbf{85.33} & 49.33          & 70.00          & 80.00          & 62.00                   \\
 \midrule
\multirow{5}{*}{Oracle}         & \multirow{5}{*}{Pair+Dream Machine} & 2+1    & 11                            & 5.71                 & 5.84                 & 81.67          & 93.67          & 95.67          & 72.33          & 90.00          & 96.33          & 80.08                   \\
                                &                                & 2+3      & 11              & 5.80                 & 5.00                 & 81.33          & 94.33          & 95.00          & 73.33          & 91.00          & 96.67          & 81.19                   \\
                                &                                & 2+8  & 11                             & 6.81                 & 6.00                 & 81.33          & 91.67          & 94.00          & 71.67          & 87.67          & 96.00          & 78.20                   \\
                                &                                & 2+38   & 11                            & 7.42                 & 7.10                  & 78.33          & 91.33          & 93.67          & 65.33          & 84.33          & 94.67          & 75.26                   \\
                                &                                & 2+114 & 1                            & 9.21                 & 9.68                 & 74.33          & 89.33          & 92.67          & 59.67          & 77.33          & 90.67          & 70.70                  
\\ \bottomrule
\end{tabular}
}
\caption{\textbf{Ablation study on the number of input images to the pose estimator on ScanNet dataset.} "\# Images" denotes the total number of images provided to the DUSt3R pose estimator, where 2 images are from the original pair and the remaining images are sampled from the generated video. Using 5 images, as used in the main paper, shows the best performance. "\# Samples" indicates the sampling iterations per video. For the experiment with 2+114 images, only one sampling was conducted instead of 11, since the video consists of 114 frames in total.}
\label{tab:ablation_frames}
\end{table*}

\subsection{Ablation study on distance metrics}
In the main paper, we quantify video inconsistency using the medoid distance $D_\text{med}$.
We also define the total distance as
\begin{equation}
    D_\text{total} = D_\text{med} + \text{dist}\left(\hat{T}_\text{med}, f_\text{pose}(\{I_A, I_B\})\right),
\end{equation}
where $\hat{T}_\text{med}$ is the medoid relative pose, and $f_\text{pose}(\{I_A, I_B\})$ is the pose estimated from the original image pair.
We select the video with the lowest $D_\text{total}$ and output the predicted medoid relative pose $\hat{T}_\text{med}$ as the consensus pose.

In Table~\ref{tab:ablation_distance}, we present an ablation study on the distance metrics by comparing predictions based on $D_\text{total}$, $D_\text{med}$, and $D_\text{bias}$, where
\begin{equation}
    D_\text{bias} = \text{dist}\left(\hat{T}_\text{med}, f_\text{pose}(\{I_A, I_B\})\right).
\end{equation}
Our results indicate that for most datasets and video models, both $D_\text{total}$ and $D_\text{med}$ obtain comparable results and consistently outperform the DUSt3R baseline, which only takes original image pairs. 
However, on the Cambridge Landmarks dataset using Dream Machine as the generative video model, utilizing $D_\text{med}$ alone results in a significant increase in rotation error from $11.96^\circ$ to $19.37^\circ$ compared to using $D_\text{total}$. 
This demonstrates that incorporating $D_\text{bias}$ into the distance metric enhances robustness and generalization ability across different datasets and video models.

\subsection{Ablation study on the number of input images}

The oracle showing the tendency as worse performance when using more video frames, which is likely due to less randomless in sampling, and also video might contain inconsistent content, which might degenerate the performance if the original input pair is less considered in pose estimation and post-optimization process.

We present an ablation study on the number of input images to the pose estimator in Table~\ref{tab:ablation_frames}. 
The baseline DUSt3R takes only the original image pair as input, utilizing two images. To explore the impact of varying the number of input frames, we conducted experiments with 3, 5, 10, 40, and 116 images. These configurations correspond to sampling 1, 3, 8, 38, and 114 frames from the video generated by Dream Machine, respectively. Since the Dream Machine video consists of 114 frames in total, the configuration with 116 images involves sampling all frames once, while the other configurations involve multiple sampling iterations (11 times for all except the 116-image setup).

The results indicate that using five images, as adopted in the main paper, yields the best performance across most metrics, including Mean Rotation Error (MRE), Mean Translation Error (MTE), and AUC$_{30\degree}$. 
In addition, the oracle results reveal a trend of degenerating performance as the number of video frames increases. This decline is likely due to reduced randomness in sampling and the less-emphasisis on the original input pair during the pose estimation and post-optimization processes. 
Overall, these results indicate that using five frames provides a robust and generalizable approach, avoiding the pitfalls associated with both insufficient and excessive frame counts.

\begin{table*}[tb]
\centering
\resizebox{0.87\textwidth}{!}{
\begin{tabular}{llllccccccccc}
\toprule
\multirow{2}{*}{Yaw range} & \multirow{2}{*}{\# Pairs} & \multirow{2}{*}{Pose estimator} & \multirow{2}{*}{Input data} & \multirow{2}{*}{MRE$\downarrow$} & \multirow{2}{*}{MTE$\downarrow$} & \multicolumn{3}{c}{R$_\text{acc}\uparrow$}                        & \multicolumn{3}{c}{t$_\text{acc}\uparrow$}                     & \multirow{2}{*}{AUC$_{\text{30°}}$$\uparrow$} \\
 \cmidrule(lr){7-9} \cmidrule(lr){10-12}
                           &                                 &                                 &                             &                      &                      & 5°              & 15°             & 30°           & 5°              & 15°             & 30°          &                         \\
\midrule
\multirow{3}{*}{[0°, 50°]}   & \multirow{3}{*}{200}            & DUSt3R                          & Pair                        & 11.33                & 22.50                & 76.00          & 83.50          & 89.00          & 43.50          & 67.50          & 78.00          & 60.20                   \\
                           &                                 & Ours                    & Dream Machine               & \textbf{9.12}        & \textbf{15.75}       & \textbf{78.00} & \textbf{87.50} & \textbf{90.00} & \textbf{45.00} & \textbf{70.50} & \textbf{82.00} & \textbf{61.82}          \\
                           &                                 & Oracle                          & Dream Machine               & 2.72                 & 4.77                 & 88.50          & 97.50          & 99.00          & 71.50          & 91.50          & 97.50          & 84.52                   \\
\midrule
\multirow{3}{*}{[0°, 25°]}   & \multirow{3}{*}{100}            & DUSt3R                          & Pair                        & 9.29                 & 20.76                & 82.00          & 87.00          & \textbf{91.00} & 41.00          & 74.00          & 80.00          & 61.97                   \\
                           &                                 & Ours                    & Dream Machine               & \textbf{8.41}        & \textbf{15.30}       & \textbf{85.00} & \textbf{89.00} & 89.00          & \textbf{44.00} & \textbf{75.00} & \textbf{82.00} & \textbf{62.43}          \\
                           &                                 & Oracle                          & Dream Machine               & 2.12                 & 4.77                 & 91.00          & 99.00          & 99.00          & 74.00          & 93.00          & 96.00          & 85.20                   \\
\midrule
\multirow{3}{*}{[25°, 50°]}  & \multirow{3}{*}{100}            & DUSt3R                          & Pair                        & 13.36                & 24.25                & 70.00          & 80.00          & 87.00          & \textbf{46.00} & 61.00          & 76.00          & 58.43                   \\
                           &                                 & Ours                    & Dream Machine               & \textbf{9.83}        & \textbf{16.20}       & \textbf{71.00} & \textbf{86.00} & \textbf{91.00} & \textbf{46.00} & \textbf{66.00} & \textbf{82.00} & \textbf{61.20}          \\
                           &                                 & Oracle                          & Dream Machine               & 3.33                 & 4.78                 & 86.00          & 96.00          & 99.00          & 69.00          & 90.00          & 99.00          & 83.83                  
\\ \bottomrule
\end{tabular}
}
\caption{\textbf{Camera pose estimation results on large overlapping pairs with yaw changes in the range [0°, 50°] on the ScanNet dataset.} Our method demonstrates improved performance over DUSt3R on input pairs alone, in scenarios with significant overlapping regions.}
\label{tab:supp_scannet_yaw1}
\end{table*}

\begin{table*}[tb]
\centering
\resizebox{0.87\textwidth}{!}{
\begin{tabular}{llllccccccccc}
\toprule
\multirow{2}{*}{Yaw range} & \multirow{2}{*}{\# Pairs} & \multirow{2}{*}{Pose estimator} & \multirow{2}{*}{Input data} & \multirow{2}{*}{MRE$\downarrow$} & \multirow{2}{*}{MTE$\downarrow$} & \multicolumn{3}{c}{R$_\text{acc}\uparrow$}                        & \multicolumn{3}{c}{t$_\text{acc}\uparrow$}                     & \multirow{2}{*}{AUC$_{\text{30°}}$$\uparrow$} \\
 \cmidrule(lr){7-9} \cmidrule(lr){10-12}
                           &                                 &                                 &                             &                      &                      & 5°              & 15°             & 30°           & 5°              & 15°             & 30°          &                         \\
\midrule
\multirow{3}{*}{[65°, 180°]}  & \multirow{3}{*}{200}            & DUSt3R                          & Pair                        & \textbf{83.48}                & 58.93                & \textbf{20.50} & 28.50          & 31.50          & 19.00          & 26.00          & 31.50          & 20.88                   \\
                            &                                 & Ours                    & Dream Machine               & 83.94       & \textbf{37.81}       & 18.50          & \textbf{30.00} & \textbf{33.00} & \textbf{20.00} & \textbf{31.00} & \textbf{44.00} & \textbf{21.28}          \\
                            &                                 & Oracle                          & Dream Machine               & 36.94                & 11.91                & 38.00          & 51.50          & 57.00          & 41.50          & 71.50          & 89.00          & 39.50                   \\
\midrule
\multirow{3}{*}{[65°, 110°]}  & \multirow{3}{*}{95}             & DUSt3R                          & Pair                        & 59.24                & 50.44                & \textbf{31.58} & 44.21          & 46.32          & \textbf{25.26} & 35.79          & 43.16          & 31.30                   \\
                            &                                 & Ours                    & Dream Machine               & \textbf{56.35}       & \textbf{33.16}       & 28.42          & \textbf{48.42} & \textbf{50.53} & 24.21          & \textbf{38.95} & \textbf{51.58} & \textbf{32.53}          \\
                            &                                 & Oracle                          & Dream Machine               & 15.98                & 11.30                & 60.00          & 75.79          & 77.89          & 46.32          & 70.53          & 89.47          & 56.11                   \\
\midrule
\multirow{3}{*}{[110°, 180°]} & \multirow{3}{*}{105}            & DUSt3R                          & Pair                        & \textbf{105.41}      & 66.61                & \textbf{10.48} & \textbf{14.29} & \textbf{18.10} & 13.33          & 17.14          & 20.95          & \textbf{11.46}          \\
                            &                                 & Ours                    & Dream Machine               & 108.89               & \textbf{42.02}       & 9.52           & 13.33          & 17.14          & \textbf{16.19} & \textbf{23.81} & \textbf{37.14} & 11.11                   \\
                            &                                 & Oracle                          & Dream Machine               & 55.91                & 12.46                & 18.10          & 29.52          & 38.10          & 37.14          & 72.38          & 88.57          & 24.48                  
\\ \bottomrule
\end{tabular}
}
\caption{\textbf{Camera pose estimation results on non-overlapping pairs with yaw changes in the range [65°, 180°] on the ScanNet dataset.} The performance of DUSt3R and our method significantly drops in this challenging non-overlapping scenario. While our method obtains better translation estimation, it exhibits slightly worse rotation estimation compared to DUSt3R.}
\label{tab:supp_scannet_yaw2}
\end{table*}

\begin{table*}[tb]
\centering
\resizebox{0.87\textwidth}{!}{
\begin{tabular}{llllccccccccc}
\toprule
\multirow{2}{*}{Yaw range} & \multirow{2}{*}{\# Pairs} & \multirow{2}{*}{Pose estimator} & \multirow{2}{*}{Input data} & \multirow{2}{*}{MRE$\downarrow$} & \multirow{2}{*}{MTE$\downarrow$} & \multicolumn{3}{c}{R$_\text{acc}\uparrow$}                        & \multicolumn{3}{c}{t$_\text{acc}\uparrow$}                     & \multirow{2}{*}{AUC$_{\text{30°}}$$\uparrow$} \\
 \cmidrule(lr){7-9} \cmidrule(lr){10-12}
                           &                                 &                                 &                             &                      &                      & 5°              & 15°             & 30°           & 5°              & 15°             & 30°          &                         \\
\midrule
\multirow{3}{*}{[0°, 50°]}   & \multirow{3}{*}{200}            & DUSt3R                          & Pair                        & 4.28                 & 11.04                & 79.00          & 95.50          & 98.00           & 49.00          & 89.00          & 93.00          & 73.60                   \\
                           &                                 & Ours                     & Dream Machine               & \textbf{3.24}        & \textbf{8.16}        & \textbf{81.50} & \textbf{97.50} & \textbf{99.50}  & \textbf{49.00} & \textbf{90.50} & \textbf{96.00} & \textbf{76.17}          \\
                           &                                 & Oracle                          & Dream Machine               & 1.68                 & 3.16                 & 93.00          & 100.00         & 100.00          & 85.50          & 98.00          & 99.00          & 89.80                   \\
                           \midrule
\multirow{3}{*}{[0°, 25°]}   & \multirow{3}{*}{67}             & DUSt3R                          & Pair                        & 2.78                 & 10.87                & 91.73          & 96.99          & 98.50           & 42.86          & 87.22          & 93.23          & 73.31                   \\
                           &                                 & Ours                     & Dream Machine               & \textbf{1.87}        & \textbf{8.48}        & \textbf{94.74} & \textbf{99.25} & \textbf{100.00} & \textbf{45.86} & \textbf{88.72} & \textbf{96.24} & \textbf{76.17}          \\
                           &                                 & Oracle                          & Dream Machine               & 1.04                 & 3.62                 & 98.50          & 100.00         & 100.00          & 81.95          & 97.74          & 98.50          & 89.82                   \\
                           \midrule
\multirow{3}{*}{[25°, 50°]}  & \multirow{3}{*}{133}            & DUSt3R                          & Pair                        & 7.25                 & 11.37                & 53.73          & 92.54          & 97.01           & \textbf{61.19} & 92.54          & 92.54          & 74.18                   \\
                           &                                 & Ours                     & Dream Machine               & \textbf{5.96}        & \textbf{7.54}        & \textbf{55.22} & \textbf{94.03} & \textbf{98.51}  & 55.22          & \textbf{94.03} & \textbf{95.52} & \textbf{76.17}          \\
                           &                                 & Oracle                          & Dream Machine               & 2.95                 & 2.26                 & 82.09          & 100.00         & 100.00          & 92.54          & 98.51          & 100.00         & 89.75                  
\\ \bottomrule
\end{tabular}
}
\caption{\textbf{Camera pose estimation results on large overlapping pairs with yaw changes in the range [0°, 50°] on DL3DV-10K.} DUSt3R already performs strongly on this center-facing dataset, and Our method still achieves slight improvements over DUSt3R.}
\label{tab:supp_scannet_dl3dv_yaw1}
\end{table*}

\begin{table*}[tb]
\centering
\resizebox{0.87\textwidth}{!}{
\begin{tabular}{llllccccccccc}
\toprule
\multirow{2}{*}{Yaw range} & \multirow{2}{*}{\# Pairs} & \multirow{2}{*}{Pose estimator} & \multirow{2}{*}{Input data} & \multirow{2}{*}{MRE$\downarrow$} & \multirow{2}{*}{MTE$\downarrow$} & \multicolumn{3}{c}{R$_\text{acc}\uparrow$}                        & \multicolumn{3}{c}{t$_\text{acc}\uparrow$}                     & \multirow{2}{*}{AUC$_{\text{30°}}$$\uparrow$} \\
 \cmidrule(lr){7-9} \cmidrule(lr){10-12}
                           &                                 &                                 &                             &                      &                      & 5°              & 15°             & 30°           & 5°              & 15°             & 30°          &                         \\
\midrule
\multirow{3}{*}{[90°, 180°]}  & \multirow{3}{*}{200}            & DUSt3R                          & Pair                        & 19.20                & 15.00                & \textbf{32.50} & 79.00          & 85.50          & 52.50          & 86.00          & 87.00          & 65.07                   \\
                            &                                 & Ours                     & Dream Machine               & \textbf{16.06}       & \textbf{9.62}        & 31.50          & \textbf{82.00} & \textbf{89.50} & \textbf{53.50} & \textbf{88.50} & \textbf{91.50} & \textbf{66.37}          \\
                            &                                 & Oracle                          & Dream Machine               & 8.18                 & 3.77                 & 68.00          & 92.00          & 95.00          & 86.00          & 96.50          & 97.50          & 82.18                   \\ \midrule
\multirow{3}{*}{[90°, 110°]}  & \multirow{3}{*}{158}            & DUSt3R                          & Pair                        & 17.81                & 14.73                & \textbf{30.38} & 79.11          & 86.71          & 48.73          & 87.34          & 87.97          & 64.64                   \\
                            &                                 & Ours                     & Dream Machine               & \textbf{14.66}       & \textbf{9.17}        & 28.48          & \textbf{82.28} & \textbf{91.14} & \textbf{50.63} & \textbf{89.87} & \textbf{93.04} & \textbf{66.20}          \\
                            &                                 & Oracle                          & Dream Machine               & 6.35                 & 3.11                 & 67.09          & 93.67          & 96.84          & 86.08          & 97.47          & 98.73          & 83.14                   \\ \midrule
\multirow{3}{*}{[110°, 180°]} & \multirow{3}{*}{42}             & DUSt3R                          & Pair                        & 24.42                & 15.99                & 40.48          & 78.57          & 80.95          & \textbf{66.67} & 80.95          & 83.33          & 66.67                   \\
                            &                                 & Ours                     & Dream Machine               & \textbf{21.31}       & \textbf{11.30}       & \textbf{42.86} & \textbf{80.95} & \textbf{83.33} & 64.29          & \textbf{83.33} & \textbf{85.71} & \textbf{66.98}          \\
                            &                                 & Oracle                          & Dream Machine               & 15.05                & 6.26                 & 71.43          & 85.71          & 88.10          & 85.71          & 92.86          & 92.86          & 78.57                  
\\ \bottomrule
\end{tabular}
}
\caption{\textbf{Camera pose estimation results on pairs with large yaw changes in the range [90°,180°] on DL3DV-10K.} The center-facing nature of this dataset ensures overlapping regions despite significant viewpoint changes, enabling DUSt3R to produce reasonable estimations. Our method obtains better pose estimation results over DUSt3R.}
\label{tab:supp_scannet_dl3dv_yaw2}
\end{table*}

\begin{table*}[t]
\scriptsize
\centering
\setlength{\tabcolsep}{3pt} %
\begin{tabular}{llccccccccccccccc}
\toprule
                                &                             & \multicolumn{5}{c}{Cambridge Landmarks}                       &  & \multicolumn{9}{c}{ScanNet}                                                                                                                                  \\

                             \cmidrule(lr){3-7} \cmidrule(lr){9-17}
\multirow{2}{*}{Pose estimator} & \multirow{2}{*}{Input data} & \multirow{2}{*}{MRE$\downarrow$} & \multicolumn{3}{c}{R$_\text{acc}\uparrow$} & \multirow{2}{*}{AUC$_{\text{30}}\uparrow$}                        &  & \multirow{2}{*}{MRE$\downarrow$} & \multirow{2}{*}{MTE$\downarrow$} & \multicolumn{3}{c}{R$_\text{acc}\uparrow$}                        & \multicolumn{3}{c}{t$_\text{acc}\uparrow$}                  & \multirow{2}{*}{AUC$_{\text{30}}\uparrow$} \\
\cmidrule(lr){4-6} \cmidrule(lr){11-13} \cmidrule(lr){14-16}
                                &                             &                      & 5°              & 15°             & 30°         &    &  &                      &                      & 5°              & 15°             & 30°            & 5°              & 15°          & 30°          &                         \\
                                \midrule
SIFT+N.N.                     & \multirow{4}{*}{Pair}       & 97.64                & 15.17          & 22.41          & 24.48          & 20.49                   &  & 112.95               & 48.99                & 2.06           & 3.44           & 5.50           & 23.02          & 25.09          & 31.62          & 1.82                    \\
LOFTR                           &                             & 30.30                & 31.38          & 56.55          & 70.00          & 51.63                   &  & 64.46                & 45.49                & 8.33           & 17.00          & 22.00          & 27.00          & 28.33          & 35.33          & 6.43                    \\
DUSt3R                          &                             & 13.28                & 63.45          & 87.24          & 88.97          & 77.23                   &  & 21.31                & 24.72                & 65.33          & 76.33          & 79.00          & 48.33          & 68.33          & 73.67          & 60.34                   \\
MASt3R                          &                             & 36.55                & 28.62          & 64.83          & 74.14          & 55.69                   &  & 24.35                & 17.93                & 44.00          & 73.33          & 79.67          & 38.00          & 67.33          & 77.67          & 55.10                   \\
\midrule
\multirow{3}{*}{Ours (DUSt3R)}  & DynamiCrafter               & 12.70                   & \cellcolor{red}65.17    & 88.97                   & 90.34                   & \cellcolor{orange}79.00    &  & \cellcolor{orange}18.96 & 16.42                   & \cellcolor{orange}68.00 & \cellcolor{red}82.33    & \cellcolor{orange}84.33 & \cellcolor{orange}48.67 & \cellcolor{orange}71.67 & 80.33                   & \cellcolor{orange}62.14 \\
                                & Runway                      & \cellcolor{red}10.78    & \cellcolor{orange}64.83 & \cellcolor{red}91.03    & \cellcolor{red}94.14    & \cellcolor{red}80.59    &  & 19.93                   & 16.31                   & 67.67                   & \cellcolor{orange}81.33 & \cellcolor{orange}84.33 & \cellcolor{red}51.00    & \cellcolor{red}72.33    & 80.67                   & 61.83                   \\
                                & Dream Machine                & \cellcolor{orange}11.96 & 57.93                   & \cellcolor{orange}89.66 & \cellcolor{orange}92.76 & 78.67                   &  & \cellcolor{red}17.65    & 15.88                   & \cellcolor{red}68.67    & \cellcolor{orange}81.33 & \cellcolor{red}85.33    & 47.67                   & 71.33                   & \cellcolor{orange}82.33 & \cellcolor{red}63.06    \\
\midrule
\multirow{3}{*}{Ours (MASt3R)}  & DynamiCrafter               & 31.43                   & 34.83                   & 70.00                   & 76.55                   & 60.03                   &  & 21.97                   & 16.48                   & 53.00                   & 75.67                   & 80.00                   & 40.67                   & 70.33                   & 80.00                   & 57.90                   \\
                                & Runway                      & 29.04                   & 42.07                   & 72.76                   & 78.97                   & 63.57                   &  & 21.68                   & \cellcolor{orange}15.28 & 50.33                   & 75.67                   & 81.67                   & 41.00                   & 70.00                   & \cellcolor{red}83.33    & 57.19                   \\
                                & Dream Machine                & 27.47                   & 34.48                   & 74.14                   & 80.69                   & 63.14                   &  & 19.91                   & \cellcolor{red}15.05    & 53.00                   & 78.67                   & 83.00                   & 41.00                   & 70.33                   & \cellcolor{orange}82.33 & 58.28                   \\
\midrule
Oracle & All Video Models & 3.65                 & 90.69          & 96.55          & 98.28          & 92.08                   &  & 5.80                 & 5.00                 & 81.33          & 94.33          & 95.00          & 73.33          & 91.00          & 96.67          & 81.19  
\\ \bottomrule
\end{tabular}
\caption{\textbf{Camera pose estimation results on outward-facing datasets (Cambridge Landmarks and ScanNet).} We evaluate the pairwise pose estimation task using our method based on two pose estimators DUSt3R and MASt3R. Our method consistently outperforms both DUSt3R and MASt3R when using input pairs alone across three video generators. We also present an Oracle baseline that selects the best possible relative pose recovered from all generated videos.}
\label{tab:mast3r_outward}
\end{table*}

\begin{table*}[t]
\scriptsize
\setlength{\tabcolsep}{3pt} %
\resizebox{\textwidth}{!}{
\begin{tabular}{llcccccccccccccccccc}
\toprule
                                &                             & \multicolumn{9}{c}{DL3DV-10K}                                                                                                                                                & \multicolumn{9}{c}{NAVI}                                                                                                                                     \\
                                \cmidrule(lr){3-11} \cmidrule(lr){12-20}
\multirow{2}{*}{Pose estimator} & \multirow{2}{*}{Input data} & \multirow{2}{*}{MRE$\downarrow$} & \multirow{2}{*}{MTE$\downarrow$} & \multicolumn{3}{c}{R$_\text{acc}\uparrow$}                        & \multicolumn{3}{c}{t$_\text{acc}\uparrow$}                     & \multirow{2}{*}{AUC$_{\text{30°}}$$\uparrow$} &         \multirow{2}{*}{MRE$\downarrow$} & \multirow{2}{*}{MTE$\downarrow$} & \multicolumn{3}{c}{R$_\text{acc}\uparrow$}                        & \multicolumn{3}{c}{t$_\text{acc}\uparrow$}                  & \multirow{2}{*}{AUC$_{\text{30°}}$$\uparrow$} \\ 
\cmidrule(lr){5-7} \cmidrule(lr){8-10} \cmidrule(lr){14-16} \cmidrule(lr){17-19}
                                &                             &                      &                      & 5°              & 15°             & 30°           & 5°              & 15°             & 30°          &                                 &                      &                      & 5°              & 15°             & 30°             & 5°           & 15°          & 30°             &                         \\ \midrule
SIFT+N.N.                     & \multirow{4}{*}{Pair}       & 76.64                & 46.80                & 18.06          & 28.09          & 33.44          & 31.77          & 33.11          & 36.45          & 12.11                   & 107.46               & 45.10                & 4.67           & 6.67           & 7.33           & 16.33          & 17.00          & 19.00          & 3.20                    \\
LOFTR                           &                             & 35.92                & 41.76                & 37.67          & 52.33          & 61.00          & 40.00          & 41.00          & 45.33          & 23.53                   & 71.34                & 51.21                & 6.67           & 14.33          & 19.00          & 24.67          & 25.33          & 29.33          & 4.88                    \\
DUSt3R                          &                             & 10.72                & 13.08                & 39.67          & 87.33          & 94.00          & 55.33          & 83.67          & 89.00          & 66.99                   & 8.65                 & 7.88                 & 68.67          & 92.67          & 94.67          & 69.00          & 92.33          & 95.00          & 78.66                   \\
MASt3R                          &                             & \cellcolor{red}4.13    & \cellcolor{red}3.88    & \cellcolor{red}83.67    & 98.00                   & \cellcolor{red}99.33    & \cellcolor{red}88.33    & 95.33                   & 97.00                   & \cellcolor{red}87.22    & 5.59                   & \cellcolor{orange}5.23 & \cellcolor{orange}71.67 & 94.33                   & 98.00                   & 69.67                   & \cellcolor{orange}96.00 & 98.00                   & 80.84                   \\
\midrule

\multirow{3}{*}{Ours (DUSt3R)} & DynamiCrafter               & 10.02                  & 9.13                   & 38.33                   & 87.33                   & 95.67                   & 58.33                   & 87.00                   & 93.00                   & 67.97                   & 8.26                   & 6.57                   & 68.00                   & 92.67                   & 95.67                   & 69.00                   & 91.67                   & 96.67                   & 78.78                   \\
                                & Runway                      & 9.49                   & 8.81                   & 41.33                   & 90.33                   & \cellcolor{orange}96.67 & 57.33                   & 86.67                   & 92.33                   & 69.44                   & 8.08                   & 6.24                   & 67.67                   & 93.67                   & 96.00                   & 67.67                   & 93.33                   & 97.00                   & 79.02                   \\
                                & Dream Machine                & 9.13                   & 8.72                   & 41.33                   & 90.33                   & 96.33                   & 57.67                   & 86.33                   & 94.67                   & 69.11                   & 7.85                   & 6.51                   & 69.33                   & 93.67                   & 95.33                   & \cellcolor{red}71.00    & 93.00                   & 95.67                   & 79.06                   \\
\midrule
\multirow{3}{*}{Ours (MASt3R)} & DynamiCrafter               & 4.49                   & 4.04                   & 81.33                   & \cellcolor{orange}98.67 & \cellcolor{red}99.33    & 86.33                   & \cellcolor{orange}95.67 & \cellcolor{red}97.67    & 85.86                   & \cellcolor{orange}5.29 & 5.61                   & 69.00                   & \cellcolor{orange}96.67 & \cellcolor{red}98.67    & 63.00                   & 95.67                   & \cellcolor{red}98.67    & 80.21                   \\
                                & Runway                      & \cellcolor{orange}4.17 & \cellcolor{orange}4.01 & \cellcolor{orange}81.67 & \cellcolor{red}99.00    & \cellcolor{red}99.33    & \cellcolor{orange}87.33 & \cellcolor{red}96.00    & \cellcolor{orange}97.33 & \cellcolor{orange}86.79 & \cellcolor{red}5.28    & \cellcolor{red}5.20    & \cellcolor{red}72.67    & 96.33                   & \cellcolor{red}98.67    & 69.00                   & \cellcolor{red}97.00    & \cellcolor{red}98.67    & \cellcolor{red}81.63    \\
                                & Dream Machine                & 4.30                   & 4.21                   & 80.67                   & \cellcolor{red}99.00    & \cellcolor{red}99.33    & 85.33                   & 94.67                   & 97.00                   & 85.88                   & 5.66                   & 5.45                   & 70.00                   & \cellcolor{red}97.33    & \cellcolor{orange}98.33 & \cellcolor{orange}70.00 & \cellcolor{orange}96.00 & \cellcolor{orange}98.33 & \cellcolor{orange}81.42 \\
\midrule
Oracle & All Video Models & 1.35                 & 1.05                 & 97.67          & 100.00         & 100.00         & 96.33          & 99.33          & 100.00         & 95.83                   & 2.23                 & 1.67                 & 94.33          & 99.33          & 100.00         & 94.33          & 100.00         & 100.00         & 92.90                  
\\ \bottomrule
\end{tabular}
}
\caption{\textbf{Camera pose estimation results on center-facing datasets (DL3DV-10K and NAVI).} MASt3R demonstrates significantly improved performance on these center-facing datasets compared to outward-facing ones. We evalute our method based on two pose estimators DUSt3R and MASt3R. Our method obtains comparable results on the DL3DV-10K dataset and slightly better performance on the NAVI dataset, demonstrating that using a video model does not hinder performance even when DUSt3R and MASt3R are already strong.}
\label{tab:mast3r_inward}
\end{table*}

\end{document}